\documentclass{article} 
\usepackage{iclr2027_conference,times}

\usepackage{amsmath,amsfonts,bm}

\def\eqref#1{equation~\ref{#1}}

\def\1{\bm{1}}

\DeclareMathAlphabet{\mathsfit}{\encodingdefault}{\sfdefault}{m}{sl}
\SetMathAlphabet{\mathsfit}{bold}{\encodingdefault}{\sfdefault}{bx}{n}

\newcommand{\E}{\mathbb{E}}

\newcommand{\R}{\mathbb{R}}

\newcommand{\StreamIndex}{T}
\newcommand{\SequenceLength}{L_{\mathrm{seq}}}
\newcommand{\MotionFrameIndex}{f}

\newcommand{\MemoryLag}{i}

\newcommand{\SpeechHidden}{h}
\newcommand{\SpeechDim}{d_h}
\newcommand{\RunAheadHorizon}{k}
\newcommand{\FutureOffset}{j}
\newcommand{\ProvisionalHidden}{\widehat{\SpeechHidden}}
\newcommand{\SpeechWindow}{\mathcal{W}}
\newcommand{\OnlineSpeechWindow}{\widehat{\SpeechWindow}}
\newcommand{\ProbeRSquared}{R^2}
\newcommand{\SpeechProjection}{W_{\SpeechHidden}}
\newcommand{\SpeechParams}{\psi}
\newcommand{\SpeechBackbone}{F_{\SpeechParams}}
\newcommand{\TemporalTransformer}{F_{\SpeechParams}^{H}}
\newcommand{\DepthTransformer}{F_{\SpeechParams}^{d}}
\newcommand{\BatchedRunAhead}{\mathcal{B}_{\SpeechParams}}
\newcommand{\TokenBuffer}{\mathbf{T}}
\newcommand{\UpdatedTokenBuffer}{\TokenBuffer^{\prime}}
\newcommand{\UserAudioTokens}[1]{u_{#1}}
\newcommand{\PredictedUserAudioTokens}[1]{\hat{u}_{#1}}
\newcommand{\AvatarTokens}[1]{a_{#1}}
\newcommand{\SpeechCache}{\mathcal{KV}}

\newcommand{\HeadEyeGroup}{\mathrm{he}}
\newcommand{\ExpressionJawGroup}{\mathrm{ej}}
\newcommand{\MotionGroupIndex}{g}
\newcommand{\IdentityMatrix}{I}
\newcommand{\MotionDim}{d_x}
\newcommand{\QueueLength}{Q}
\newcommand{\ActiveFrameCount}{N}
\newcommand{\CleanMotion}{\mathbf{x}}
\newcommand{\MotionNoise}{\boldsymbol\epsilon}
\newcommand{\NoisyMotion}{\mathbf{z}}
\newcommand{\MotionWidth}{d}
\newcommand{\FlowTime}{u}
\newcommand{\FlowTimes}{\mathbf{\FlowTime}}

\newcommand{\MotionParams}{\theta}
\newcommand{\VelocitySymbol}{v}
\newcommand{\VelocityField}{\VelocitySymbol_{\MotionParams}}
\newcommand{\PredictedVelocity}{\widehat{\mathbf{\VelocitySymbol}}}
\newcommand{\FrameVelocity}[1]{\VelocitySymbol_{\MotionParams,#1}}
\newcommand{\MotionConditions}{\mathcal{C}}

\newcommand{\SharedSteps}{K}
\newcommand{\ExtraSteps}{K_{\mathrm{add}}}
\newcommand{\ExpressionJawSteps}{K_{\ExpressionJawGroup}}
\newcommand{\ExpressionJawStepSize}{\Delta_{\ExpressionJawGroup}}
\newcommand{\HeadEyeStepSize}{\Delta_{\HeadEyeGroup}}
\newcommand{\GroupStepSize}{\Delta_{\MotionGroupIndex}}

\newcommand{\BehaviorSet}{\mathcal{Y}}
\newcommand{\DialogueHidden}[1]{\SpeechHidden_{#1}}
\newcommand{\ContextMemory}[1]{\mathbf{H}_{#1}}
\newcommand{\ContextLength}{L}
\newcommand{\SemanticContext}[1]{\mathbf{c}^{\mathrm{sem}}_{#1}}
\newcommand{\BehaviorProb}[1]{\mathbf{p}^{\mathrm{beh}}_{#1}}
\newcommand{\StateProb}[1]{\mathbf{p}^{\mathrm{state}}_{#1}}
\newcommand{\RouterParams}{\phi}
\newcommand{\RouterHead}{g_{\RouterParams}}
\newcommand{\AcceptedBehavior}[1]{b_{#1}}
\newcommand{\DialogueState}[1]{s_{#1}}
\newcommand{\StyleRouter}{\mathcal{R}}
\newcommand{\MotionProjection}{\Pi_{\mathrm{motion}}}
\newcommand{\FadeWeight}[1]{\alpha_{#1}}
\newcommand{\PreviousStyleTokens}[1]{\mathbf{P}^{\mathrm{prev}}_{#1}}
\newcommand{\StyleTokens}[1]{\mathbf{P}^{\mathrm{style}}_{#1}}
\newcommand{\BlendedStyleTokens}[1]{\widetilde{\mathbf{P}}^{\mathrm{style}}_{#1}}
\newcommand{\PersonaTokens}{\mathbf{P}^{\mathrm{per}}}

\newcommand{\MotionToken}[1]{\mathbf{X}_{#1}}
\newcommand{\GesturePhase}[1]{\tau_{#1}}
\newcommand{\BehaviorEmbedding}{E_{\mathrm{beh}}}
\newcommand{\StateEmbedding}{E_{\mathrm{state}}}
\newcommand{\PhaseEmbedding}{E_{\mathrm{phase}}}
\newcommand{\TotalBehaviorLoss}{\mathcal{L}}
\newcommand{\MotionFlowLoss}{\mathcal{L}_{\mathrm{FM}}}
\newcommand{\RoutingLoss}{\mathcal{L}_{\mathrm{route}}}
\newcommand{\RoutingWeight}{\lambda_{\mathrm{route}}}

\usepackage{hyperref}
\usepackage{url}
\usepackage{kotex}
\usepackage{graphicx}
\usepackage{fontawesome5}
\usepackage{booktabs,tabularx}
\usepackage{xcolor,colortbl}
\usepackage{enumitem}
\usepackage{multirow}
\usepackage{algorithm}
\usepackage{algorithmicx}
\usepackage{algpseudocode}
\usepackage{amssymb,multirow,nicematrix,tikz}
\usetikzlibrary{patterns}
\title{FacePlex: Toward Natural Full-Duplex \\  Conversational Avatars}

\author{%
Habin Lim\textsuperscript{1}\footnotemark[1]\quad
Hah Min Lew\textsuperscript{1}\footnotemark[1]\quad
Jae-Ho Lee\textsuperscript{1}\footnotemark[1]\quad
Min-Jae Kim\textsuperscript{1} \\
\bfseries Seungeun Lee\textsuperscript{2}\quad
Ji-Su Kang\textsuperscript{2}\quad
Gyeong-Moon Park\textsuperscript{1}\footnotemark[2] \\
\textsuperscript{1}Korea University, \textsuperscript{2}Klleon}

\newcommand{\para}[1]{%
    \par\noindent\textbf{#1}\hspace{0.5em}%
}

\iclrarxivcopy 
\begin{document}

\maketitle

\begin{abstract}
Natural human conversation is inherently a real-time interaction in which speech and facial behavior continuously evolve.
Enabling such interaction requires a conversational avatar to jointly generate speech and facial motion in real time, prepare facial motion for upcoming speech before the corresponding audio is emitted, and produce non-verbal reactions that reflect the ongoing dialogue.
However, existing conversational avatar systems cannot address such requirements: audio-driven methods rely on pre-given speech, while joint streaming generation alone does not ensure anticipatory articulation or semantically appropriate reactions.
We propose \textbf{FacePlex}, a unified framework for full-duplex speech--facial motion generation and real-time avatar rendering.
FacePlex jointly coordinates speech, facial motion, and Gaussian splatting rendering on a shared streaming timeline.
To prepare facial motion for upcoming speech, FacePlex predicts a short speech continuation and uses its future hidden states through asymmetric conditioning and denoising, while continuously updating them as new user audio arrives without observing future user input.
For dialogue-grounded non-verbal behavior, we construct \textit{SemReact}, a dataset aligning dialogue context, reaction semantics, and facial motion, and introduce a semantic behavior router that guides continuous motion during both speaking and listening.
Extensive experiments show improved audio-visual synchronization and facial articulation, natural dialogue-grounded non-verbal responses, and low-latency of 122 ms end-to-end avatar interaction.
\end{abstract}

{\renewcommand*{\thefootnote}{}%
  \footnotetext{
    \textbf{\textsuperscript{$\ast$}Equal contribution}
    \quad
    \textbf{\textsuperscript{$\dagger$}Corresponding author}
    \quad
    \faGithub \ Project page: \href{https://hahminlew.github.io/faceplex/}{https://hahminlew.github.io/faceplex}
  }%
}
\setcounter{footnote}{0}

\begingroup
\ificlrarxiv
  \setlength{\parskip}{5pt}
\fi

\section{Introduction}
\label{sec:introduction}

Conventional conversational AI systems typically follow a turn-based interaction pattern: the model waits until the user provides a full utterance, and then responds, and the cycle repeats.
Natural human conversation, however, extends beyond discrete turn-taking: speech and facial behavior occur throughout the exchange rather than only at turn boundaries~\citep{levinson2016turn,agrawal2025seamless}.
For example, listeners provide verbal backchannels, nods, and facial expressions while another person is speaking~\citep{knudsen2020forgotten,ng2022learning}, and speakers prepare facial articulation for upcoming sounds and use non-verbal behavior to convey their stance.
A conversational avatar must therefore do more than animate emitted speech: it must continuously generate speech and facial motion, anticipate upcoming articulation, and produce non-verbal behavior grounded in the dialogue.
We formulate these capabilities as three requirements for natural conversational avatars.

\textbf{(R1) Continuous Full-Duplex Speech--Motion Streaming.}
The system must continue receiving user input while concurrently generating avatar speech and facial motion, without waiting for complete utterances.
Here, \emph{full-duplex} denotes concurrent processing of incoming signals and generation of outgoing responses at each timestep~\citep{defossez2024moshi,roy2026personaplex}.
Because human conversation combines rapid responses~\citep{stivers2009universals} with tightly synchronized vocal and visual signals~\citep{holler2019multimodal}, speech generation, motion synthesis, and avatar rendering must remain temporally aligned within a shared end-to-end latency budget.

\textbf{(R2) Anticipatory Facial Articulation.}
Facial motion must reflect speech that is being prepared before acoustic emission.
Classical articulatory studies show that facial articulation begins before the corresponding acoustic event~\citep{benguerel1974coarticulation,abry1995modeling}.
Visible mouth movements can precede voice onset by approximately 100--300\,ms~\citep{chandrasekaran2009natural}, while anticipatory labial coarticulation may emerge more than 200\,ms before the acoustic emission of a target rounded vowel~\citep{lo2023language}.
Natural articulation therefore requires access to upcoming speech information early enough to guide facial motion while preserving causal streaming, without buffering complete utterances or accessing future user observations.

\textbf{(R3) Dialogue-Grounded Non-Verbal Behavior.}
Non-verbal facial behavior must reflect the semantic context of the ongoing dialogue.
Listeners' nods can express affiliation with a speaker's stance~\citep{stivers2008stance}, while head shakes can convey negation independently or together with speech~\citep{kendon2002some}.
This requires supervision that connects dialogue context to both an appropriate reaction and its continuous facial-motion realization, beyond speaking/listening labels alone.

\begin{figure}[t]
    \centering
    \includegraphics[width=\linewidth]{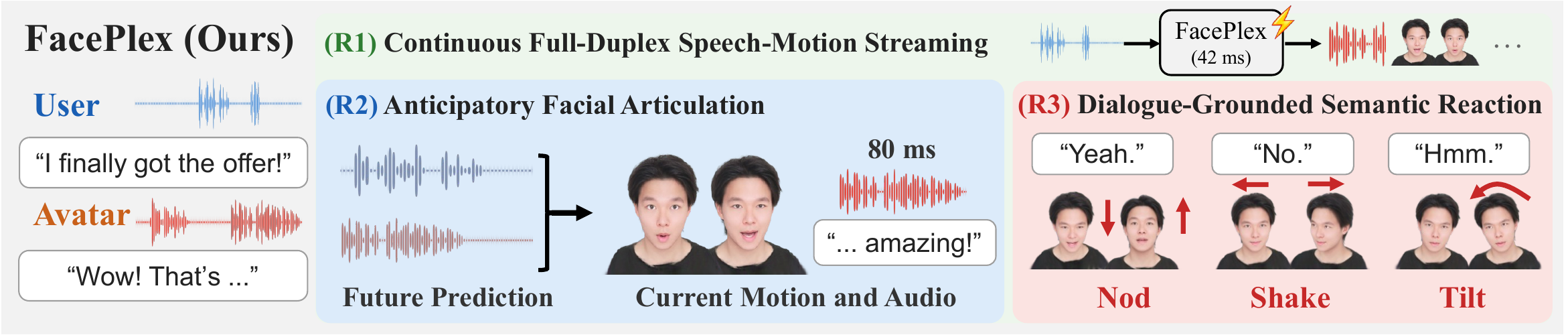}
    \caption{FacePlex targets three requirements for natural conversational avatars: \textbf{(R1)} continuous full-duplex speech--motion streaming, \textbf{(R2)} anticipatory facial articulation using predicted future context, and \textbf{(R3)} dialogue-grounded semantic reactions such as nodding, head shaking, and head tilting. Speech and facial motion are coordinated in 80\,ms streaming steps.}
    \label{fig:motivation}
\end{figure}

However, existing speech-driven facial-generation paradigms address only subsets of these requirements (Figure~\ref{fig:motivation}).
Conventional utterance-conditioned methods assume that the full utterance speech is already available~\citep{fan2022faceformer,xing2023codetalker,sun2024diffposetalk}, generating only facial motion rather than jointly producing speech and motion online.
Methods that model both speaking and listening, such as UniLS~\citep{chu2026unils} and DualTalk~\citep{peng2025dualtalk}, similarly animate conversations from complete utterance audio.
More recently, Moshi-Face~\citep{jiang2026integrating} supports full-duplex speech and face generation at the token level, but its non-causal face codec leaves fully causal visual streaming unresolved.
Existing paradigms therefore address only subsets of \textbf{(R1)--(R3)}, leaving their joint realization within a single causal streaming avatar unresolved.

To address these requirements, we propose \textbf{FacePlex}, a unified framework for full-duplex joint speech--facial motion generation and real-time avatar rendering.
For \textbf{(R1)}, FacePlex jointly generates speech and facial motion on a shared full-duplex streaming timeline while continuously processing user input. 
Gaussian splatting (GS) rendering~\citep{kerbl20233d} is integrated into this pipeline for real-time visual output.
For \textbf{(R2)}, we first analyze the temporal dependence between conversational hidden states and facial motion and find that nearby future hidden states provide stronger cues for current articulation than past or temporally aligned states alone.
Motivated by this observation, we generate a short speech continuation, keep its avatar tokens fixed, and recompute its hidden states as new user audio arrives.
To meet the 80\,ms causal streaming budget, we perform this recomputation with \textit{batched run-ahead prediction} for motion conditioning.
We couple these updated hidden states with \textit{asymmetric conditioning and denoising}: expression and jaw use the latest current and future hidden states just before output, while head and eye motion are progressively denoised across successive windows.
For \textbf{(R3)}, we construct \textit{SemReact}, a conversational reaction dataset that combines augmented real motion from CCDb-HG~\citep{vuillecard2024ccdb} with synthetic semantic control to align dialogue context, reaction semantics, and facial motion.
Using this supervision, a \emph{semantic behavior router} selects reactions from past conversational hidden states and conditions the motion generator through \emph{style token routing}, enabling dialogue-grounded non-verbal behavior during both speaking and listening while retaining the avatar's motion characteristics.

To evaluate whether these designs satisfy the three requirements, we conduct comprehensive experiments, demonstrating improved audio-visual synchronization and facial articulation in offline evaluation, naturalistic non-verbal responses grounded in dialogue context, and low-latency end-to-end operation.
Together, these results demonstrate the continuous full-duplex speech--motion streaming, anticipatory facial articulation, and dialogue-grounded non-verbal behavior within a unified conversational avatar. Our contributions are summarized as follows:

\begin{itemize}[leftmargin=10pt,itemsep=0pt,parsep=0pt,topsep=1pt]
    \item We introduce \textbf{FacePlex}, a unified conversational avatar framework that enables continuous full-duplex speech--motion streaming, anticipatory facial articulation, and dialogue-grounded non-verbal behavior.
    
    \item Based on our temporal analysis, we combine \textit{batched run-ahead prediction} with \textit{asymmetric conditioning and denoising} to enable anticipatory facial articulation within a causal streaming budget.
    
    \item We construct \textit{SemReact}, a dataset aligning dialogue context, reaction semantics, and facial motion, enabling our \textit{semantic behavior router} with \textit{style token routing} to generate dialogue-grounded non-verbal behavior during both speaking and listening.
    
    \item Extensive experiments demonstrate state-of-the-art facial-motion performance across speaking and listening in offline evaluation, while achieving the highest user-study ratings, with 122 ms, end-to-end latency.
\end{itemize}

\section{Related Work}
\label{sec:related_work}
\para{Full-Duplex Speech and Audio-Visual Conversation.}
Full-duplex speech models process incoming speech while generating responses, enabling overlap and interruptions~\citep{defossez2024moshi,wang2025freezeomni,zhang2025omniflatten,roy2026personaplex}.
Moshi-Face~\citep{jiang2026integrating} jointly generates speech and face tokens with a temporally causal dialogue model.
Its Face Transformer's non-causal attention operates across tokens within each timestep and does not itself violate temporal causality.
The limitation lies in its non-causal face codec: the authors explicitly identify a causal codec as future work needed for fully streaming visual input and output.
This distinguishes token-level generation from a causal visual pipeline, without implying that the codec necessarily buffers entire utterances.
FacePlex integrates speech, continuous facial motion, and rendering while predicting upcoming content solely from available observations.

\para{Audio-Driven Facial Motion Generation.}
Prior works animate speech using learned styles~\citep{cudeiro2019capture}, transformers~\citep{fan2022faceformer}, discrete motion priors~\citep{xing2023codetalker}, autoregression~\citep{chu2025artalk}, and diffusion~\citep{sun2024diffposetalk}, or directly synthesize talking-head video~\citep{prajwal2020lip,zhang2023sadtalker}.
These methods assume that the speech to be animated is available, whereas joint online generation must obtain articulatory cues before emitting its own speech.
Articulatory studies motivate this distinction by showing that lip movements can precede the corresponding sounds~\citep{benguerel1974coarticulation,chandrasekaran2009natural,lo2023language}.
FacePlex supports anticipatory facial articulation by combining run-ahead prediction with asymmetric conditioning and denoising, using upcoming hidden states without access to future observations.

\para{Conversational Reactions and Semantic Behavior Control.}
Listener-behavior generation methods model responses to a conversational partner~\citep{ng2022learning,ng2023can,zhou2022responsive,luo2024reactface}, while DualTalk~\citep{peng2025dualtalk} and UniLS~\citep{chu2026unils} model both speaking and listening.
However, conversational roles alone do not capture how non-verbal responses vary with dialogue context.
STEER~\citep{teotia2026steer} supports explicit behavior control and causal motion generation with Gaussian avatar rendering, but requires externally supplied controls and leaves their selection from dialogue content to future work.
FacePlex uses SemReact's dialogue--reaction--motion supervision to train a semantic behavior router, selecting reactions from conversational context and realizing them as continuous non-verbal motion through style token routing.

\section{FacePlex}
\label{sec:method}


\begin{figure*}[t]
    \centering
    \includegraphics[width=\textwidth]{Figures/MainFigure.pdf}
    \caption{Overview of FacePlex.
    Sequential self-replay generates future speech; batched run-ahead $\BatchedRunAhead$ then recomputes the hidden states as new user audio arrives.
    A semantic behavior router with style token routing selects context-dependent non-verbal behavior.
    At step $\StreamIndex$, FacePlex prepares speech and facial motion for step $\StreamIndex+1$, with facial motion rendered using Gaussian splatting.}
    \label{fig:faceplex_overview}
\end{figure*}

FacePlex generates speech and facial motion while continuously listening to the user (Figure~\ref{fig:faceplex_overview}).
It combines anticipatory facial articulation from future hidden states with dialogue-grounded non-verbal behavior, while maintaining synchronized speech and motion on a shared streaming timeline.
We use PersonaPlex-7B~\citep{roy2026personaplex} as the speech backbone $\SpeechBackbone$, with frozen parameters $\SpeechParams$.
It consists of a temporal transformer $\TemporalTransformer$ and a depth transformer $\DepthTransformer$.
At each 80\,ms processing step $\StreamIndex$, $\TemporalTransformer$ computes $\DialogueHidden{\StreamIndex+1}\in\R^{\SpeechDim}$ from token inputs through position $\StreamIndex$, with $\SpeechDim=4096$.
The hidden-state index refers to the next output position, so $\DialogueHidden{\StreamIndex+1}$ is available at step $\StreamIndex$.
$\DepthTransformer$ uses a hidden state to generate predicted user audio tokens and avatar audio tokens.

Our motion generator produces a pair of FLAME~\citep{FLAME:SiggraphAsia2017} frames per step, giving 25\,Hz animation.
At step $\StreamIndex$, it prepares the motion pair for output at $\StreamIndex+1$, synchronized with the corresponding avatar audio.
The motion generator uses asymmetric conditioning and denoising with hidden states from run-ahead prediction.
It also uses history, persona, and style tokens.
History tokens represent previously completed motion frames.
Persona tokens encode a fixed reference for the avatar's motion characteristics.
Style tokens encode a reaction reference selected by the semantic behavior router at each step.
The following subsections focus on our temporal analysis, batched run-ahead prediction, asymmetric conditioning and denoising, semantic behavior routing, and streaming-optimized avatar rendering.

\subsection{When does Audio Affect Motion?}
\label{sec:future_context}

Facial articulation often begins before the acoustic emission of the corresponding speech sound~\citep{benguerel1974coarticulation,abry1995modeling,lo2023language}.
Studies comparing the timing of mouth movements and speech acoustics report that mouth movements can begin about 100--300\,ms before voice emission~\citep{chandrasekaran2009natural}.
These findings raise a practical question for streaming motion generation: \textbf{\emph{Is past and current speech information sufficient to generate accurate facial motion, or is information about future speech also needed?}}

To investigate this question, we extract hidden states from Seamless Interaction conversations~\citep{agrawal2025seamless} and
probe how well a single state at each time offset predicts the target motion by measuring the coefficient of determination ($\ProbeRSquared$).
For jaw and expression, $\ProbeRSquared$ peaks for hidden states 160\,ms ahead of the target motion and decreases farther into the future.
Head and eye motion show low $\ProbeRSquared$ across offsets (Figure~\ref{fig:future_context_analysis}(a)).
Appendix~\ref{app:hidden_state_probe} describes the probing setup in detail.

We set a window of four hidden states and evaluate its temporal placement by training motion generators with windows of the same length at different offsets.
Shifting the window into the near future reduces Speaking Lip Vertex Error (S-LVE), but shifting it further into the future provides no further benefit (Figure~\ref{fig:future_context_analysis}(b)).
For the motion pair to be output at $\StreamIndex+1$, we therefore use the state at that target position and the next $\RunAheadHorizon=3$ states: $\SpeechWindow_{\StreamIndex}=[\DialogueHidden{\StreamIndex+1},\ldots,\DialogueHidden{\StreamIndex+\RunAheadHorizon+1}]$.
Guided by this analysis, we adopt asymmetric conditioning and denoising schedules tailored to the distinct temporal dependencies of different motion components (Figure~\ref{fig:future_context_analysis}(c)).
Section~\ref{sec:asymmetric_refinement} describes these in detail.

\subsection{Batched Run-Ahead Prediction for Motion Conditioning}
\label{sec:run_ahead}

As shown by our analysis in Section~\ref{sec:future_context}, motion generation benefits from future hidden states.
Computing these states requires user audio that has not yet arrived.
Therefore, we predict the future user audio using PersonaPlex, which is trained to predict both user audio tokens ($\PredictedUserAudioTokens{\StreamIndex}$) and avatar audio tokens ($\AvatarTokens{\StreamIndex}$)~\citep{defossez2024moshi,roy2026personaplex}.
We feed the predicted user audio tokens and avatar audio tokens back into the model to generate tokens for the next step, a process called \emph{self-replay}: $(\AvatarTokens{\StreamIndex+1},\PredictedUserAudioTokens{\StreamIndex+1})=\DepthTransformer(\TemporalTransformer(\AvatarTokens{\StreamIndex},\PredictedUserAudioTokens{\StreamIndex}))$.
Repeating this process produces future avatar audio and the hidden states needed for motion generation without waiting for future user audio.

\begin{figure}
    \centering
    \includegraphics[width=1\linewidth]{Figures/Analysis.pdf}
    \caption{
    \textbf{(a)} Coefficient of determination ($\ProbeRSquared$) for motion predictions from a single hidden state at various time offsets.
    \textbf{(b)} Speaking Lip Vertex Error (S-LVE) across four-state conditioning windows.
    \textbf{(c)} Asymmetric conditioning and denoising schedules for each motion component.}
    \label{fig:future_context_analysis}
\end{figure}

However, repeating this sequential process whenever new user audio arrives would require multiple sequential backbone forward passes per streaming step,
making it difficult to meet the 80\,ms computation budget for each step.
We therefore use batched run-ahead to recompute the hidden-state window while keeping the buffered avatar audio tokens fixed.
New predicted user audio tokens and avatar audio tokens are generated only at the end of the buffer.
We initialize the token buffer $\TokenBuffer$ once at the start of each conversation using sequential self-replay.
Starting with the initial avatar audio tokens $\AvatarTokens{0}$ and received user audio tokens $\UserAudioTokens{0}$, this warm-up fills four buffer positions: $\TokenBuffer_0=[(\AvatarTokens{0},\UserAudioTokens{0}),(\AvatarTokens{1},\PredictedUserAudioTokens{1}),(\AvatarTokens{2},\PredictedUserAudioTokens{2}),(\AvatarTokens{3},\PredictedUserAudioTokens{3})]$ (Figure~\ref{fig:faceplex_overview}).
The buffered avatar audio tokens determine the speech that the user will hear.
Appendix~\ref{app:run_ahead_consistency} evaluates the consistency between committed audio and recomputed hidden states, along with their effects on speech predictions.

At subsequent streaming steps $\StreamIndex\ge1$, the buffer is $\TokenBuffer_{\StreamIndex}=[(\AvatarTokens{\StreamIndex},\PredictedUserAudioTokens{\StreamIndex}),\ldots,(\AvatarTokens{\StreamIndex+\RunAheadHorizon},\PredictedUserAudioTokens{\StreamIndex+\RunAheadHorizon})]$.
When user audio for step $\StreamIndex$ arrives, we replace the predicted user audio tokens $\PredictedUserAudioTokens{\StreamIndex}$ with the received user audio tokens $\UserAudioTokens{\StreamIndex}$, resulting in $\UpdatedTokenBuffer_{\StreamIndex}=[(\AvatarTokens{\StreamIndex},\UserAudioTokens{\StreamIndex}),\ldots,(\AvatarTokens{\StreamIndex+\RunAheadHorizon},\PredictedUserAudioTokens{\StreamIndex+\RunAheadHorizon})]$.
Since $\UpdatedTokenBuffer_{\StreamIndex}$ already contains all input tokens, batched run-ahead can recompute all hidden states in the window in a single causal forward pass through $\TemporalTransformer$.
Through causal attention, the received user audio also changes the hidden states at later positions, even though their input tokens remain fixed.
We denote this batched run-ahead function by $\BatchedRunAhead$, which uses the frozen speech backbone $\SpeechBackbone$.
$\BatchedRunAhead$ outputs the hidden state window, predicted user audio tokens, and avatar audio tokens:
\begin{equation}
    (\OnlineSpeechWindow_{\StreamIndex},\PredictedUserAudioTokens{\StreamIndex+\RunAheadHorizon+1},\AvatarTokens{\StreamIndex+\RunAheadHorizon+1})
    =\BatchedRunAhead(\UpdatedTokenBuffer_{\StreamIndex}).
\end{equation}
Here, $\OnlineSpeechWindow_{\StreamIndex}$ approximates $\SpeechWindow_{\StreamIndex}$, with $\ProvisionalHidden$ denoting hidden states computed using predicted user audio tokens at future positions.
For $\StreamIndex\ge1$, batched run-ahead shifts and recomputes $\OnlineSpeechWindow_{\StreamIndex-1}$ as $\OnlineSpeechWindow_{\StreamIndex}$ using the received user audio tokens $\UserAudioTokens{\StreamIndex}$.
\begin{equation}
    \OnlineSpeechWindow_{\StreamIndex-1}=[\DialogueHidden{\StreamIndex},\ProvisionalHidden_{\StreamIndex+1\mid\StreamIndex-1},\ldots,\ProvisionalHidden_{\StreamIndex+\RunAheadHorizon\mid\StreamIndex-1}]
    \xrightarrow{\UserAudioTokens{\StreamIndex}}
    \OnlineSpeechWindow_{\StreamIndex}=[\DialogueHidden{\StreamIndex+1},\ProvisionalHidden_{\StreamIndex+2\mid\StreamIndex},\ldots,\ProvisionalHidden_{\StreamIndex+\RunAheadHorizon+1\mid\StreamIndex}].
\end{equation}
Only the last hidden state $\ProvisionalHidden_{\StreamIndex+\RunAheadHorizon+1\mid\StreamIndex}$ is passed to the depth transformer $\DepthTransformer$ to generate new predicted user audio tokens and avatar audio tokens.
We append the new tokens and remove the oldest token pair to form $\TokenBuffer_{\StreamIndex+1}$.
At $\StreamIndex=0$, we use $\UpdatedTokenBuffer_0=\TokenBuffer_0$ for the first batched run-ahead step, which generates $\PredictedUserAudioTokens{4}$ and $\AvatarTokens{4}$ and forms $\TokenBuffer_1$ in the same way.
The avatar audio tokens $\AvatarTokens{\StreamIndex+1}$ for the next output remain fixed in the buffer while their corresponding motion pair is prepared. (See Appendix~\ref{app:run_ahead_details}.)

\subsection{Asymmetric Conditioning and Denoising}
\label{sec:asymmetric_refinement}
FacePlex controls the temporal context of each FLAME component through asymmetric denoising schedules.
Each component is denoised at selected streaming steps, using the hidden state window $\OnlineSpeechWindow_{\StreamIndex}$ recomputed by $\BatchedRunAhead$ at step $\StreamIndex$.
These schedules build on rolling flow matching~\citep{ergasti2025rflav}, which combines the sliding-window denoising of rolling diffusion~\citep{ruhe2024rolling} with flow matching~\citep{lipman2022flow} to gradually transform noise into motion.

At step $\StreamIndex$, the motion buffer $\NoisyMotion$ holds $\QueueLength=\RunAheadHorizon+1=4$ motion pairs for output positions $\StreamIndex+1,\ldots,\StreamIndex+\RunAheadHorizon+1$, matching those covered by $\OnlineSpeechWindow_{\StreamIndex}$ (Section~\ref{sec:run_ahead}).
Each pair contains two motion tokens, corresponding to 80\,ms.
For denoising, we divide each motion token into two component groups, head--eye and expression--jaw, and apply a different schedule to each group.
A new pair is enqueued at the end of the buffer as Gaussian noise and moves one position forward per streaming step.
For the motion pair to be output at $\StreamIndex+1$, the head--eye group of each motion token receive two denoising updates at each step from $\StreamIndex-3$ to $\StreamIndex$, using the corresponding windows $\OnlineSpeechWindow_{\StreamIndex-3},\ldots,\OnlineSpeechWindow_{\StreamIndex}$.
The expression--jaw group are updated only at the final step $\StreamIndex$, using $\OnlineSpeechWindow_{\StreamIndex}$ (Figure~\ref{fig:future_context_analysis}(c)).
This final window starts at $\DialogueHidden{\StreamIndex+1}$ and covers the current and future positions relative to the target motion pair (Section~\ref{sec:future_context}).
Expression and jaw use this window as their direct speech conditioning.

At each denoising update, the motion generator with trainable parameters $\MotionParams$ predicts velocities
$\PredictedVelocity=\VelocityField(\NoisyMotion;\FlowTimes,\OnlineSpeechWindow_{\StreamIndex},\MotionConditions)$.
Here $\NoisyMotion$ contains the normalized FLAME parameters of the noisy motion buffer.
$\FlowTimes$ records the denoising progress of the head--eye and expression--jaw groups in each motion pair, from zero (noise) to one (completed motion).
$\MotionConditions$ includes the history, persona, and style tokens.
The predicted velocities $\PredictedVelocity$ are used to denoise the motion tokens according to the schedules above.
See Appendix~\ref{app:motion_updates} for the update budgets and numerical rules, and Appendix~\ref{app:motion_training} for training details.

\endgroup

\subsection{Dialogue-grounded Non-Verbal Behavior}
\label{sec:semantic_behavior}

\para{SemReact.} Grounding non-verbal motion in dialogue semantics requires supervision for both selecting an appropriate reaction and generating it as continuous motion.
We construct \textbf{SemReact}, a conversational reaction dataset combining augmented real motion from the CCDb-HG training partition~\citep{vuillecard2024ccdb} with synthetic semantic control.
Its \emph{dialogue--reaction--motion alignment} provides this supervision, enabling the higher-level behavior planning identified as future work by the externally controlled STEER~\citep{teotia2026steer}.

For the synthetic component, a language model generates dialogues with sentence-level reactions from $\BehaviorSet=\{\text{idle},\text{nod},\text{shake},\text{tilt}\}$.
A separate verification pass checks each proposed reaction against its conversational context without seeing the original label; disagreements are demoted to idle.
We synthesize both speakers' audio and blend sentence-aligned gesture primitives into continuous motion generated by UniLS~\citep{chu2026unils}.
Together with augmented real gestures, these examples link conversational context to reaction labels and motion during speaking and listening.
The synthetic corpus contains 18,075 dialogues; CCDb-HG participants are separated before constructing the real-motion bank.
Further details are provided in Appendix~\ref{app:semreact}.

\para{Semantic Behavior Router.}
A \emph{semantic behavior router} determines the avatar's speaking/listening state and reaction from recent conversational hidden states (Figure~\ref{fig:faceplex_overview}, center).
These decisions select the corresponding style tokens, and \emph{style token routing} conditions continuous motion while retaining the avatar's motion characteristics.
This connects the semantic supervision in SemReact to the rolling generator without extending its short articulation window.

At processing step $\StreamIndex$, the speaking/listening state denotes as $\DialogueState{\StreamIndex}=\arg\max_{s\in\{0,1\}}[\StateProb{\StreamIndex}]_s$.
To predict both the reaction and the speaking/listening state, we aggregate the latest $\ContextLength=25$ available hidden states, ending at $\DialogueHidden{\StreamIndex+1}$ and covering approximately 2\,s of dialogue.
Using the generator's speech projection $\SpeechProjection\in\R^{\MotionWidth\times\SpeechDim}$, we form a semantic context vector $\SemanticContext{\StreamIndex}\in\R^{\MotionWidth}$.
The router predicts reaction probabilities $\BehaviorProb{\StreamIndex}$ over $\BehaviorSet$ and speaking/listening state probabilities $\StateProb{\StreamIndex}$ over $\{0,1\}$:
\begin{equation}
    \SemanticContext{\StreamIndex}=\frac{1}{\ContextLength}\sum_{\MemoryLag=0}^{\ContextLength-1}\SpeechProjection\DialogueHidden{\StreamIndex+1-\MemoryLag},
    \qquad
    \bigl(\BehaviorProb{\StreamIndex},\StateProb{\StreamIndex}\bigr)
    =\RouterHead\!\left(\operatorname{LN}(\SemanticContext{\StreamIndex})\right).
    \label{eq:semantic_router}
\end{equation}
Here, $\MotionWidth=512$, $\MemoryLag$ is the lag in streaming steps, and $\operatorname{LN}$ is layer normalization.
The router $\RouterHead$ is an MLP with parameters $\RouterParams$ and two output heads, each applying a separate softmax over the four reaction classes or the two speaking/listening states.
Each hidden state encodes prior dialogue context, and pooling summarizes this context for both predictions.
We obtain the accepted reaction $\AcceptedBehavior{\StreamIndex}\in\BehaviorSet$ by stabilizing the highest-probability reaction class across streaming steps.

\para{Style Token Routing and Reaction Motion Generation.}
The accepted reaction $\AcceptedBehavior{\StreamIndex}$ and the router's speaking/listening state $\DialogueState{\StreamIndex}$ index short reference motion clips drawn from the training bank, with idle, nod, shake, and tilt available for both speaking and listening.
These controls and the style tokens are indexed by their processing step $\StreamIndex$ and condition the motion buffer whose first pair will be output at $\StreamIndex+1$.
The generator's shared motion projection maps the selected clip to style tokens:
\begin{equation}
    \begin{aligned}
        \StyleTokens{\StreamIndex} &= \MotionProjection\!\left(\StyleRouter(\AcceptedBehavior{\StreamIndex},\DialogueState{\StreamIndex})\right),
    \end{aligned}
    \label{eq:style_prompt_routing}
\end{equation}
where $\StyleRouter$ indexes the clip bank and $\MotionProjection$ projects 108-D motion to 512-D tokens for history and noisy motion frames.
The selected style tokens guide the current reaction, while the fixed persona tokens retain the avatar's motion characteristics.
Both form part of the motion conditioning $\MotionConditions$ in Section~\ref{sec:asymmetric_refinement}.
The generator tracks reaction progression across streaming steps, supporting coherent onset, development, and return rather than repeatedly initiating the selected gesture.

We jointly optimize $\TotalBehaviorLoss=\MotionFlowLoss+\RoutingWeight\RoutingLoss$, where $\MotionFlowLoss$ is the flow-matching loss, $\RoutingLoss$ provides cross-entropy supervision for both reaction and speaking/listening state predictions on labeled samples, and $\RoutingWeight=0.5$.
Appendix~\ref{app:semantic_behavior} details state determination, transition handling, framewise conditioning, reaction progression, and training.

\subsection{Streaming-Optimized Avatar Rendering}
\label{sec:optimization}

To realize FacePlex as an interactive visual avatar, generated motion must be rendered within the streaming budget. We use FlexAvatar~\citep{kirschstein2026flexavatar}, a generalizable avatar renderer that relies on learned neural decoding for high-fidelity synthesis. With the original FlexAvatar renderer, naive serial rendering of the two frames produced per FacePlex step consumes about 17\% of the 80\,ms budget. Profiling shows that much of this cost comes from fragmented execution around transformer attention and repeated computation that remains unchanged across frames.

We therefore reorganize renderer execution while preserving the original computation.
We introduce \emph{QKPack}, a fused operator that preserves stream-specific Query/Key normalization while directly forming the packed inputs required by joint attention~\citep{esser2024scaling} in a single GPU kernel.
We further move frame-invariant operations outside the per-frame path by pre-casting frozen linear weights from \texttt{FP32} to \texttt{FP16} and incorporating fixed upsampling modulation~\citep{karras2020analyzing} into the convolution weights. The two motion frames of each FacePlex step are rendered jointly, and the resulting fixed-shape pipeline is captured with CUDA Graphs to reduce launch overhead. Detailed formulations and equivalence analyses are provided in Appendix~\ref{app:optimization}. Together, these optimizations accelerate rendering by $3.27\times$ (Table~\ref{tab:renderer_runtime}), with negligible perceptual deviation from the original FlexAvatar renderer (LPIPS $1.04\times10^{-5}$, PSNR 68.16 dB).
\section{Experiments}
\label{sec:experiments}


\subsection{Experimental Setup}\label{sec4.1}

\noindent\textbf{Baselines.}
We compare FacePlex with ARTalk~\citep{chu2025artalk}, DualTalk~\citep{peng2025dualtalk}, UniLS~\citep{chu2026unils}, and Moshi-Face~\citep{jiang2026integrating}.
Since no public implementation of Moshi-Face was available, we adapt its method to 108-dimensional FLAME motion; implementation details and deviations from the original setting are provided in Appendix~\ref{app:moshiface_implementation}.
For avatar rendering, we use FlexAvatar~\citep{kirschstein2026flexavatar} as the rendering backbone.

\para{Evaluation Protocol.}
We evaluate on the Seamless Interaction test split~\citep{agrawal2025seamless}, following the conversational-avatar evaluation protocol of UniLS~\citep{chu2026unils}.
For speaking motion, we report Lip Vertex Error (LVE)~\citep{richard2021meshtalk}, Mean Head Distance (MHD), and Upper-Face Dynamics Deviation (FDD)~\citep{xing2023codetalker}; for listening motion, we report FDD and Fr\'echet distance (L-FID).
S-/L- denote speaking/listening intervals.

For the main motion-quality comparison, we evaluate the baselines following their default offline setting. We referred to their original papers.
We evaluate FacePlex in streaming condition.
We separately report run-ahead computation time and system latency in Section~\ref{sec:ablation}.
We measure latency from the onset of user speech to the first audible avatar response.

\subsection{Experimental Results}\label{sec4.2}
\noindent\textbf{Main Comparison.}
Table~\ref{tab:main} compares FacePlex in streaming setting and the baselines in their default offline evaluation on Seamless Interaction, alongside the user-study results in Table~\ref{tab:user}.
FacePlex achieves the lowest errors across all five reported motion-quality metrics.
FacePlex also achieves a lowest latency of 122.16\,ms.

\begin{figure}
    \centering
    \includegraphics[width=1\linewidth]{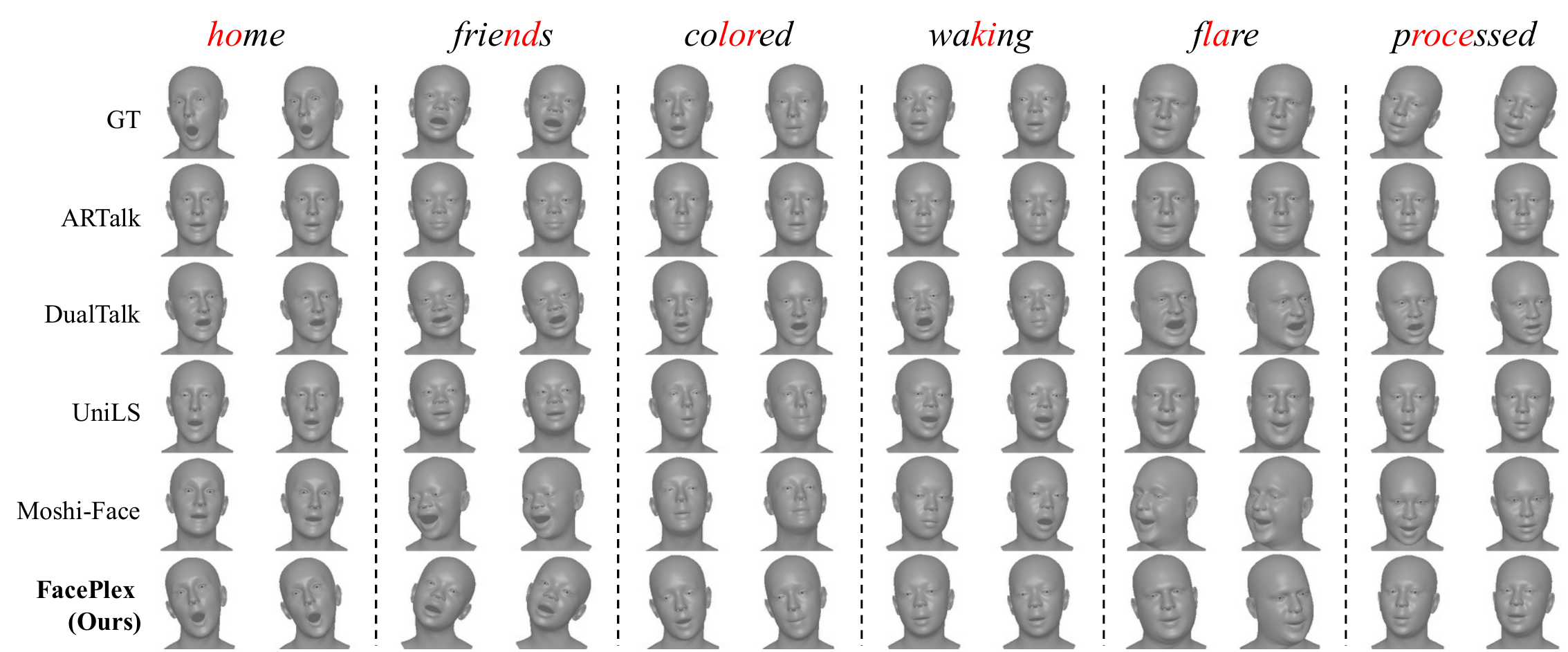}
\caption{
Qualitative comparisons.
\textbf{FacePlex} produces more expressive and speech-consistent motions, with clearer mouth articulation and more natural head movements than prior methods.
}
    \label{fig:qualitative}
\end{figure}

\noindent\textbf{Qualitative Results.} Figure~\ref{fig:qualitative} compares generated motions across different speech segments.
Compared with prior facial motion models, FacePlex produces more speech-consistent motion in these examples.
ARTalk tends to produce conservative head movements, DualTalk often shows unstable or exaggerated expressions, and UniLS is smooth but less phonetically aligned.
In contrast, FacePlex captures clearer lip motion and more natural head movements, especially for words with distinctive mouth shapes such as ``home'', ``colored'', and ``flare''. Figure~\ref{fig:nonverbal} illustrates non-verbal behaviors rendered with Gaussian splatting, including nodding, shaking, and tilting.

\begin{table*}[t]
\centering
\begin{minipage}[t]{0.66\textwidth}
    \vspace{0pt}
    \caption{Quantitative comparison on Seamless Interaction.}
    \label{tab:main}
\end{minipage}\hfill
\begin{minipage}[t]{0.32\textwidth}
    \vspace{0pt}
    \caption{User study (1--5, $\uparrow$).}
    \label{tab:user}
\end{minipage}
\par\smallskip
\begin{minipage}[t]{0.66\textwidth}
    \vspace{0pt}
    \centering
    \fontsize{6.5}{8}\selectfont
    \setlength{\tabcolsep}{0.7pt}
    \renewcommand{\arraystretch}{1.12}
    \begin{tabularx}{\linewidth}{@{}lc*{4}{r>{\fontsize{5}{6}\selectfont\raggedright\arraybackslash}X}c@{}}
        \toprule
        \textbf{Method} & Latency (s) $\downarrow$ & \multicolumn{2}{c}{S-LVE $\downarrow$} & \multicolumn{2}{c}{S-MHD $\downarrow$} & \multicolumn{2}{c}{S-FDD $\downarrow$} & \multicolumn{2}{c}{L-FDD $\downarrow$} & L-FID $\downarrow$ \\
        \midrule
        ARTalk & 4.043 & 8.385 & $\pm$2.958 & 1.906 & $\pm$0.607 & 46.878 & $\pm$19.869 & 45.006 & $\pm$20.567 & 8.811 \\
        DualTalk & 11.174 & 11.785 & $\pm$2.171 & 2.389 & $\pm$0.382 & 35.836 & $\pm$16.725 & 33.276 & $\pm$17.438 & 56.376 \\
        UniLS & 4.093 & 7.691 & $\pm$2.326 & 1.819 & $\pm$0.497 & 23.451 & $\pm$12.917 & 30.215 & $\pm$16.808 & 3.454 \\
        Moshi-Face & 4.188 & 10.605 & $\pm$2.697 & 2.457 & $\pm$0.586 & 39.307 & $\pm$21.268 & 35.595 & $\pm$20.718 & 1.924 \\
        \midrule
        \rowcolor{lightgray!40}\textbf{FacePlex (Ours)} & \textbf{0.122} & \textbf{7.162} & $\pm$1.857 & \textbf{1.674} & $\pm$0.418 & \textbf{22.839} & $\pm$13.109 & \textbf{26.623} & $\pm$15.431 & \textbf{1.407} \\
        \bottomrule
    \end{tabularx}
\end{minipage}\hfill
\begin{minipage}[t]{0.32\textwidth}
    \vspace{0pt}
\centering
\fontsize{6.5}{8}\selectfont
\setlength{\tabcolsep}{0.35pt}
\renewcommand{\arraystretch}{1.12}
\begin{tabularx}{\linewidth}{@{}l*{4}{>{\centering\arraybackslash}X}@{}}
    \toprule
    \textbf{Method} & Sync & Nat. & NVR & MOS \\
    \midrule
    ARTalk & 2.445 & 2.161 & 1.729 & 2.148 \\
    DualTalk & 3.161 & 2.665 & 2.748 & 2.948 \\
    UniLS & 2.594 & 2.510 & 2.600 & 2.626 \\
    Moshi-Face & 1.561 & 1.781 & 2.052 & 1.774 \\
    \midrule
    \rowcolor{lightgray!40}\textbf{FacePlex (Ours)} & \textbf{3.368} & \textbf{3.006} & \textbf{3.303} & \textbf{3.303} \\
    \bottomrule
\end{tabularx}

\end{minipage}
\end{table*}

\begin{figure}[t]
    \centering
    \begin{minipage}[t]{0.30\linewidth}
        \vspace{0pt}
        \fontsize{8}{9.5}\selectfont
        \setlength{\abovecaptionskip}{0pt}
        \setlength{\belowcaptionskip}{2pt}
        \setlength{\tabcolsep}{1pt}
        \renewcommand{\arraystretch}{1.464}
        \setlength{\aboverulesep}{1.5pt}
        \setlength{\belowrulesep}{1.5pt}
        {\let\normalsize\small
        \makeatletter\def\@captype{table}\makeatother
        \caption{\strut Router accuracy (\%, $\uparrow$).\newline Rule-based selects reactions from dialogue text using manual rules (Appendix~\ref{app:router_rule_baseline}).}
        \label{tab:router_rule_comparison}}
\begin{tabularx}{\linewidth}{@{}l>{\centering\arraybackslash}X>{\columncolor{lightgray!40}\centering\arraybackslash}X@{}}
    \toprule
    Reaction & Rule-based & \textbf{Ours} \\
    \midrule
    Nod & 35.87 & \textbf{75.33} \\
    Shake & 76.54 & \textbf{82.97} \\
    Tilt & 50.44 & \textbf{82.47} \\
    Mean & 54.28 & \textbf{80.26} \\
    \bottomrule
\end{tabularx}

    \end{minipage}\hfill
    \begin{minipage}[t]{0.68\linewidth}
        \vspace{0pt}
        \centering
        \includegraphics[width=\linewidth]{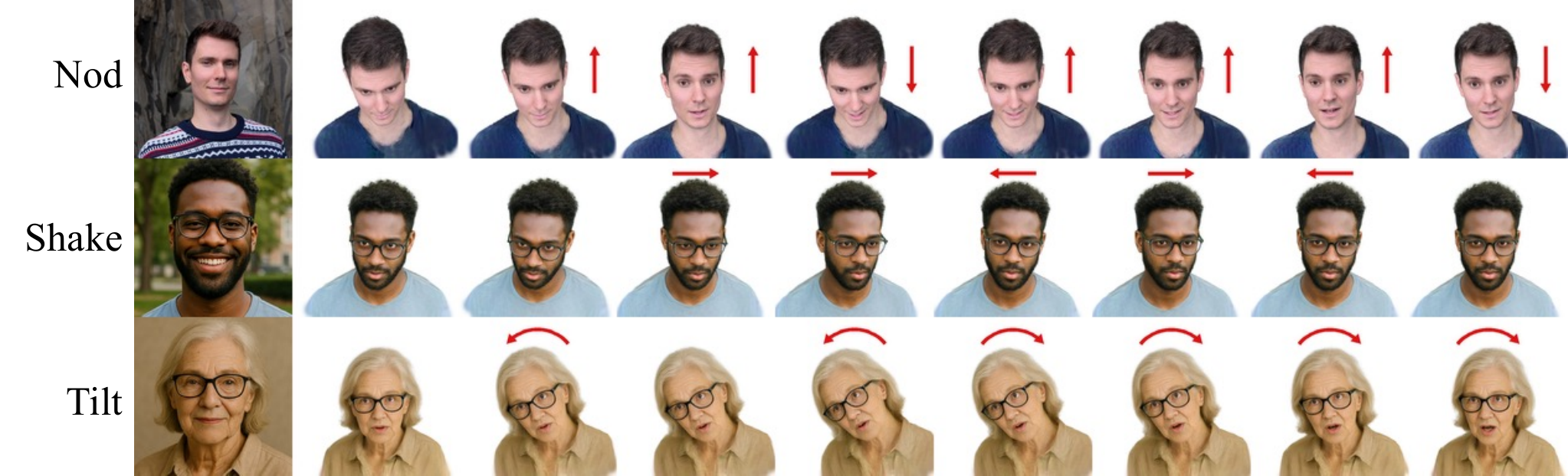}
        \setlength{\abovecaptionskip}{4pt}
        \setlength{\belowcaptionskip}{0pt}
        {\let\normalsize\small
        \caption{GS-rendered nodding, shaking, and tilting (top to bottom); arrows show head motion.}
        \label{fig:nonverbal}}
    \end{minipage}
\end{figure}

\para{User Study.}
We conducted a user study with 31 participants comparing FacePlex against ARTalk, DualTalk, UniLS, and Moshi-Face.
Each participant evaluated 5 conversation sessions.
Participants rated each video on a 1--5 scale (higher is better) for Lip-Sync (Sync), Naturalness (Nat.), Non-verbal Responsiveness (NVR), and Overall (MOS).
The interface is provided in the Appendix~\ref{app:user_study}. Table~\ref{tab:user} summarizes the perceptual comparison.
FacePlex achieves the highest mean ratings across all four criteria, with an overall score of 3.303 compared with 2.948 for the strongest baseline, DualTalk.
Its largest margin over the best baseline is in non-verbal responsiveness (3.303 vs.\ 2.748), indicating better perceived conversational semantic reactions.

\para{Video Results.} Our supplementary webpage provides qualitative video comparisons and live interactive demonstrations of FacePlex. We strongly encourage readers to view them.
\section{Ablation Studies and Further Analysis}
\label{sec:ablation}

\para{Semantic Routing.}
With ground-truth reaction intervals given, our router improves mean reaction-type accuracy over manual rules from 54.28\% to 80.26\%, particularly for nod and tilt (Table~\ref{tab:router_rule_comparison}).
This result shows that semantic behavior router provides more reliable semantic reaction selection.

\para{Component Ablations.}
Table~\ref{tab:ablation_recomputation_schedule} compares variants without hidden-state updates, asymmetric denoising, or SemReact, plus cross-attention conditioning.
Ours has the lowest S-LVE, S-MHD, and S-FDD, while listening results are mixed.
Lower L-FID without SemReact (0.759 vs.\ 1.407) may reflect differences between SemReact's augmented listening-motion distribution and Seamless Interaction; L-FID does not directly assess whether reactions are appropriate for the current dialogue.

\begingroup
\setlength{\intextsep}{2pt}
\begin{table}[!t]
    \centering
    \fontsize{8}{9.5}\selectfont
    \setlength{\abovecaptionskip}{0pt}
    \setlength{\belowcaptionskip}{2pt}
    \setlength{\tabcolsep}{0.8pt}
    \renewcommand{\arraystretch}{1.05}
    \setlength{\aboverulesep}{1.5pt}
    \setlength{\belowrulesep}{1.5pt}
    \begin{minipage}[t]{0.60\linewidth}
        \vspace{0pt}
        {\let\normalsize\small
        \caption{\strut Component ablations ($\downarrow$).}
        \label{tab:ablation_recomputation_schedule}
        \label{tab:ablation_attention}
        \label{tab:ablation_semreact_training}}
        \renewcommand{\arraystretch}{1.4}
\begin{tabularx}{\linewidth}{@{}l*{5}{>{\centering\arraybackslash}X}@{}}
    \toprule
    Configuration & S-LVE & S-MHD & S-FDD & L-FDD & L-FID \\
    \midrule
    w/o hidden state update & 7.425 & 1.716 & 23.893 & 25.999 & 1.378 \\
    w/o Asym. & 7.466 & 1.714 & 23.633 & \textbf{24.642} & 1.731 \\
    w/ Cross-attn & 7.418 & 1.721 & 24.027 & 28.728 & 1.733 \\
    w/o SemReact & 7.570 & 1.730 & 24.300 & 25.100 & \textbf{0.759} \\
    \rowcolor{lightgray!40}
    \textbf{Ours} & \textbf{7.162} & \textbf{1.674} & \textbf{22.839} & 26.623 & 1.407 \\
    \bottomrule
\end{tabularx}

    \end{minipage}\hfill
    \begin{minipage}[t]{0.39\linewidth}
        \vspace{0pt}
        {\let\normalsize\small
        \caption{\strut Renderer ablation.}
        \label{tab:renderer_runtime}}
\begin{tabularx}{\linewidth}{@{}Xccc@{}}
\toprule
\multirow{2}{*}{\textbf{Config.}}
& \textbf{Latency}
& \multirow{2}{*}{\textbf{Speedup}}
& \textbf{PSNR} \\
& (ms) $\downarrow$
&
& (dB) $\uparrow$ \\
\midrule

Original
& 6.615
& $1.00\times$
& --- \\

+ CUDA Graph
& 4.102
& $1.61\times$
& \textbf{68.63} \\

+ Frame Pair
& 3.123
& $2.12\times$
& 68.49 \\

+ QKPack
& 2.566
& $2.58\times$
& 68.16 \\

+ Folded Upsample
& 2.329
& $2.84\times$
& 68.16 \\

\rowcolor{lightgray!40}
+ Pre-casting
& \textbf{2.026}
& $\mathbf{3.27\times}$
& 68.16 \\
\bottomrule
\label{tab:renderer_ablation}
\end{tabularx}

    \end{minipage}
\end{table}

\begin{table}[!t]
    \centering
    \fontsize{8}{9.5}\selectfont
    \setlength{\abovecaptionskip}{0pt}
    \setlength{\belowcaptionskip}{2pt}
    \setlength{\tabcolsep}{0.7pt}
    \renewcommand{\arraystretch}{1.05}
    \setlength{\aboverulesep}{1.5pt}
    \setlength{\belowrulesep}{1.5pt}
    \begin{minipage}[t]{0.37\linewidth}
        \vspace{0pt}
        {\let\normalsize\small
        \caption{\strut Forward time by schedule (ms).}
        \label{tab:ablation_run_ahead_latency}}
\begin{tabularx}{\linewidth}{@{}l*{3}{c}>{\centering\arraybackslash}X@{}}
    \toprule
    Method & Wait & TF passes & LM & Total \\
    \midrule
    Full-wait & 320 & 1 & \textbf{21.19} & 341.19 \\
    Sequential & 80 & 4 & 89.39 & 169.39 \\
    \rowcolor{lightgray!40}
    \textbf{Batched (Ours)} & 80 & 1 & 25.27 & \textbf{105.27} \\
    \bottomrule
\end{tabularx}

    \end{minipage}\hfill
    \begin{minipage}[t]{0.61\linewidth}
        \vspace{0pt}
    {\let\normalsize\small
    \caption{\strut Forward time by component (ms).}
    \label{tab:hardware_latency_components}}
    \setlength{\tabcolsep}{0.5pt}
    \renewcommand{\arraystretch}{1.05}
\begin{tabular*}{\linewidth}{@{\extracolsep{\fill}}l*{10}{c}@{}}
    \toprule
    GPU & Input & Enc. & Temp. & Depth & Route & Motion & Dec. & Rend. & Fwd. & Total \\
    \midrule
    \rule[-4.5pt]{0pt}{14.9625pt}RTX 4090 & 80.00 & 3.71 & 21.71 & \textbf{9.91} & 0.37 & \textbf{6.10} & \textbf{2.31} & 4.53 & 55.13 & 135.13 \\
    \rule[-4.5pt]{0pt}{14.9625pt}H200 & 80.00 & \textbf{3.20} & \textbf{11.90} & 11.86 & \textbf{0.23} & 7.43 & 2.32 & \textbf{2.03} & \textbf{42.16} & \textbf{122.16} \\
    \bottomrule
\end{tabular*}

    \end{minipage}
\end{table}

\endgroup

\para{Renderer Optimization.}
Table~\ref{tab:renderer_runtime} cumulatively adds CUDA Graphs, joint frame-pair rendering, QKPack, folded upsampling, and weight pre-casting.
Latency decreases from 6.615 to 2.026\,ms per rendered frame ($3.27\times$ speedup), with a PSNR of 68.16\,dB relative to the original renderer output.

\para{Run-Ahead.}
For $k=3$ (Table~\ref{tab:ablation_run_ahead_latency}), full-wait incurs 320\,ms of input waiting, while sequential replay requires four transformer passes.
Batched run-ahead needs 80\,ms and one pass, reducing the readiness budget from 341.19 and 169.39 to 105.27\,ms; codec and motion processing are excluded.

\para{Streaming Runtime.}
Mean speaking forward times of 55.13\,ms (RTX 4090) and 42.16\,ms (H200) are below the 80\,ms input period (Table~\ref{tab:hardware_latency_components}), supporting streaming throughput.
Total adds 80\,ms of input acquisition to Forward (Fwd.); rendering is reported per frame and excluded from both.

Additional results and experimental protocols are provided in Appendices~\ref{app:controlled_ablation}, \ref{app:hidden_state_windows}, \ref{app:latency}, and~\ref{app:runtime_ablation}--\ref{app:renderer_equivalence}.

\section{Conclusion}
We presented FacePlex, a unified framework for full-duplex joint speech--facial motion generation and avatar rendering.
Batched run-ahead prediction makes upcoming hidden states available before the avatar's speech is emitted to the user, while asymmetric conditioning and denoising use this anticipatory context to prepare expression and jaw motion for upcoming sounds.
We also constructed SemReact, a novel dataset aligning dialogue context, reaction semantics, and facial motion, which enables a semantic behavior router to generate context-dependent reactions during both speaking and listening.
Extensive experiments and a user study demonstrate strong performance for full-duplex streaming generation of synchronized speech and expressive facial motion with only 122 ms latency.
Together, these contributions provide a foundation for conversational avatars that coordinate what they say, how they articulate, and how they respond to the meaning of an ongoing dialogue.

\clearpage

\subsection*{AI use statement}
Generative AI tools were not used to develop or refine the research methodology, research ideas, experimental design, implementation, or analysis and interpretation of results. As part of the SemReact data construction pipeline, we used Qwen2.5-14B-Instruct to generate synthetic reaction-conditioned dialogues and perform reaction verification, as described in Appendix~\ref{app:semreact_dialogues}. We also used LLMs for grammar checking and readability improvement during paper preparation. All AI-assisted content was reviewed by the authors.

\subsection*{Ethics statement}
This research adheres to the ICLR Code of Ethics. Appendix~\ref{app:limitations} discusses the potential positive and negative societal impacts of FacePlex, including applications in accessible conversational interfaces and risks related to deceptive synthetic media, impersonation, privacy, consent, and user overtrust, together with possible safeguards. Our experiments use existing conversational datasets and synthetically generated data. The user study did not require formal ethical review under our institutional guidelines; informed consent was obtained from all participants, and responses were collected anonymously. This work is intended for research on streaming multimodal generation and not for deceptive or unauthorized representations of real individuals.

\subsection*{Reproducibility statement}
We provide reproducibility details throughout the main paper and appendix. Section~\ref{sec4.1} describes the experimental setup, evaluation data, metrics, baselines, and evaluation protocol, while the appendix provides details on SemReact construction and partitions, model architecture and training, streaming generation, baseline adaptations, user-study protocols, and rendering optimizations. Appendix~\ref{app:semreact_splits} specifies the fixed data partitions and reproducibility settings for SemReact.



\bibliography{iclr2027_conference}
\bibliographystyle{iclr2027_conference}

\clearpage
\appendix
\section*{Appendix}
\newcounter{tocmain}
\newcounter{tocsub}[tocmain]

\renewcommand{\thetocmain}{\Alph{tocmain}}
\renewcommand{\thetocsub}{\thetocmain.\arabic{tocsub}}

\newcommand{\TOCMain}[2]{%
  \refstepcounter{tocmain}%
  \setcounter{tocsub}{0}%
  \noindent\hyperref[#1]{\textbf{\thetocmain. #2}} \dotfill \pageref{#1} \\
}

\newcommand{\TOCSub}[2]{%
  \refstepcounter{tocsub}%
  \noindent\hspace*{1.5em}\hyperref[#1]{\thetocsub\ #2} \dotfill \pageref{#1} \\
}

\newcommand{\TOCNew}{\vspace{0.1em}}

\setcounter{section}{0}
\setcounter{subsection}{0}
\renewcommand{\thesection}{\Alph{section}}
\renewcommand{\thesubsection}{\thesection.\arabic{subsection}}

\noindent This appendix provides data construction, implementation, training, and evaluation details, together with additional analyses of FacePlex.

\vspace{0.5em}
\noindent
\TOCMain{app:semreact}{SemReact Data Construction}
\TOCSub{app:semreact_ccdb}{CCDb-HG Real-Motion Bank}
\TOCSub{app:semreact_dialogues}{Dialogue Generation and Reaction Verification}
\TOCSub{app:semreact_audio}{Speech Synthesis and Temporal Alignment}
\TOCSub{app:semreact_motion}{Continuous Motion and Gesture Composition}
\TOCSub{app:semreact_features}{Conversational Features and Packed Representation}
\TOCSub{app:semreact_splits}{Data Partitions and Reproducibility}
\TOCSub{app:semreact_style_references}{Style Reference Bank Construction}
\TOCNew
\TOCMain{app:semantic_behavior}{Semantic Context-Guided Non-Verbal Behavior: Implementation Details}
\TOCSub{app:semantic_behavior_router}{Semantic Behavior Router: State and Reaction Decisions}
\TOCSub{app:semantic_behavior_control}{Style Token Routing and Framewise Controls}
\TOCSub{app:semantic_behavior_progression}{Reaction Progression and Temporal Continuity}
\TOCNew
\TOCMain{app:semreact_prompts}{SemReact Prompt Specifications}
\TOCSub{app:semreact_generation_prompts}{Dialogue Generation}
\TOCSub{app:semreact_prompt_slots}{Prompt Slots and Output Validation}
\TOCSub{app:semreact_verification_prompts}{Reaction Verification}
\TOCNew
\TOCMain{app:streaming_details}{Streaming Generation Details}
\TOCSub{app:motion_architecture}{Motion Generator}
\TOCSub{app:run_ahead_details}{Run-Ahead Buffer and Cache}
\TOCSub{app:run_ahead_consistency}{Consistency Between Committed Speech and Recomputed States}
\TOCSub{app:motion_updates}{Numerical Motion Updates}
\TOCSub{app:motion_training}{Motion Training}
\TOCSub{app:controlled_ablation}{Ablation Details}
\TOCSub{app:hidden_state_windows}{Hidden-State Horizon and Window Placement}
\TOCSub{app:hidden_state_probe}{Hidden-State Timing Probe}
\TOCNew
\TOCMain{app:moshiface_implementation}{Moshi-Face Baseline Implementation}
\TOCNew
\TOCMain{app:user_study}{User Study Details}
\TOCSub{appsub:protocol}{Protocol and Participants}
\TOCSub{appsub:interface}{Interface and Rating Criteria}
\TOCNew
\TOCMain{app:latency}{End-to-End Processing Latency}
\TOCNew
\TOCMain{app:optimization}{Streaming-Optimized Avatar Rendering}
\TOCSub{app:profiling}{Profiling Details}
\TOCSub{app:qkpack}{QKPack: Stream-Preserving Attention Preparation}
\TOCSub{app:static}{Removing Frame-Invariant Computation}
\TOCSub{app:cudagraph}{Fixed-Shape CUDA Graph Execution}
\TOCSub{app:runtime_ablation}{Runtime Ablation}
\TOCSub{app:renderer_equivalence}{Numerical Equivalence and Rendering Quality}
\TOCNew
\TOCMain{app:limitations}{Limitations and Broader Impact}
\TOCSub{appsub:limitations}{Limitations}
\TOCSub{appsub:pos_impact}{Potential Positive Impact}
\TOCSub{appsub:neg_impact}{Potential Negative Impact}

\vspace{1em}
\hrule
\vspace{1em}
\clearpage

\section{SemReact Data Construction}
\label{app:semreact}

\begin{figure}[htbp]
    \centering
    \includegraphics[width=\linewidth]{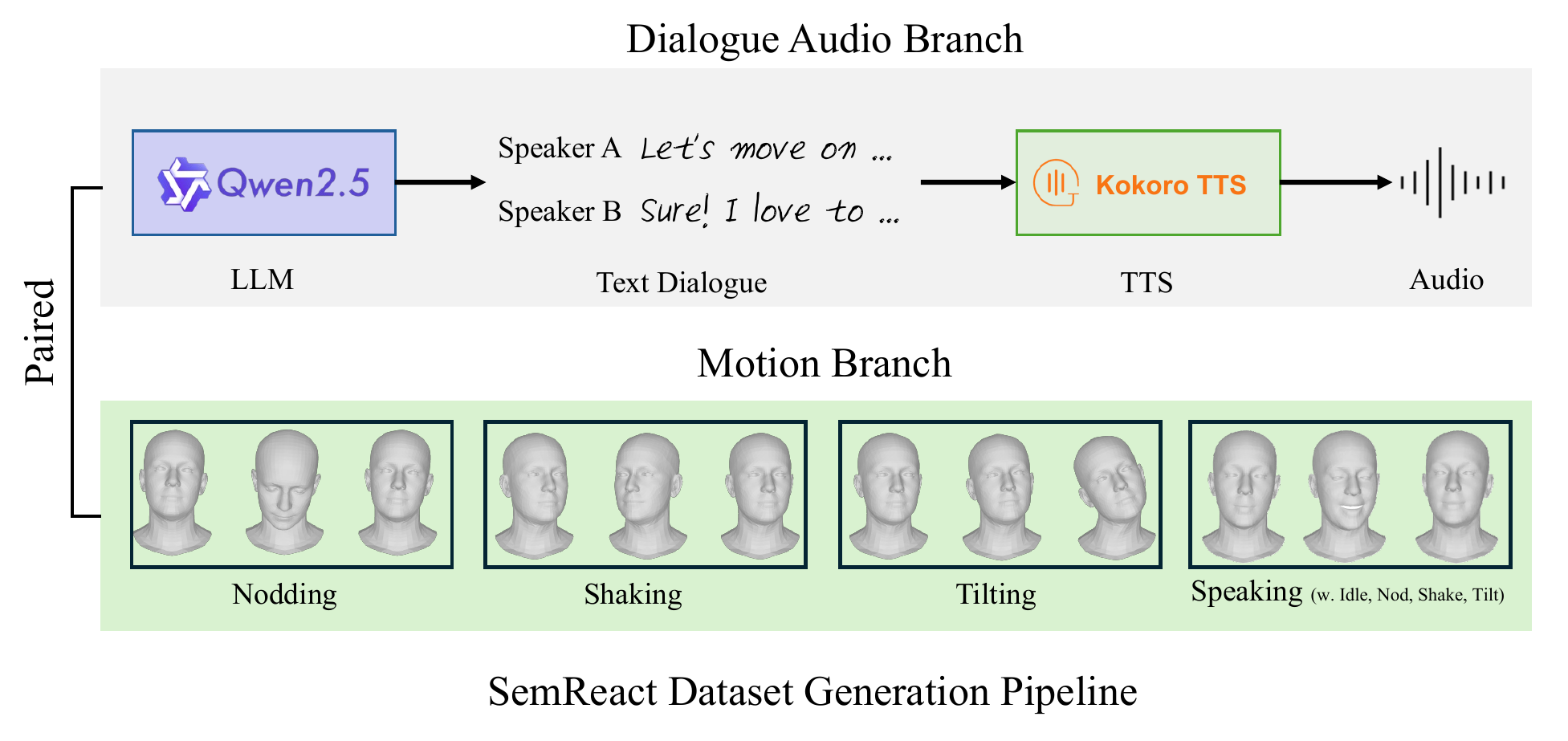}
    \caption{SemReact data generation pipeline. Qwen2.5 generates reaction-conditioned dialogues, and Kokoro synthesizes the two speakers' audio. The resulting dialogue and speech are aligned with nod, shake, and tilt reactions during speaking and listening.}
    \label{fig:semreact_pipeline}
\end{figure}

\begin{figure}[!t]
    \centering
    \includegraphics[width=\linewidth]{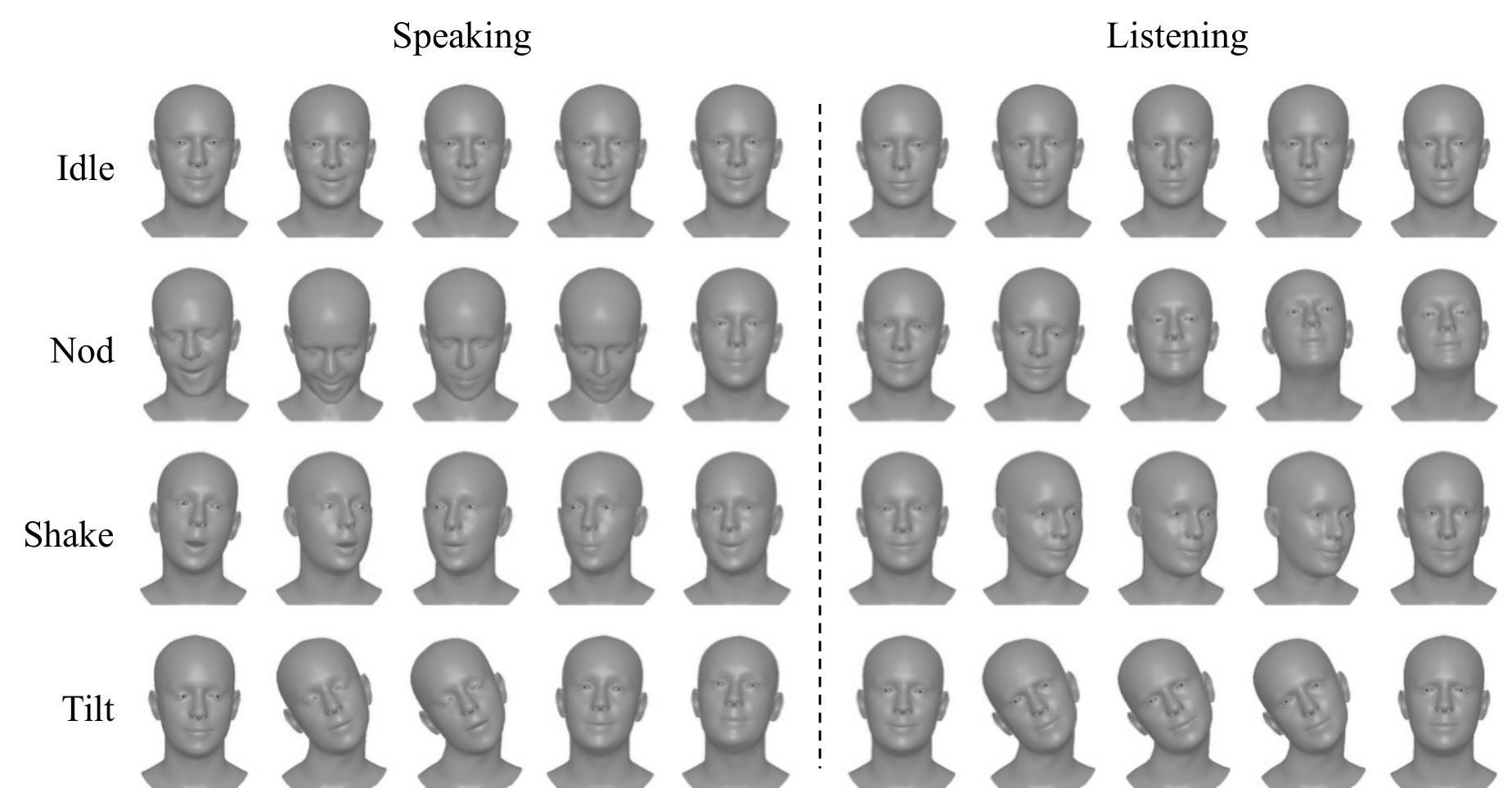}

    \medskip
    {\small
    \setlength{\tabcolsep}{4pt}
    \renewcommand{\arraystretch}{1.15}
    \begin{tabularx}{\linewidth}{@{}l>{\raggedright\arraybackslash}X>{\raggedright\arraybackslash}X@{}}
        \toprule
        \textbf{Behavior} & \textbf{Speaking} & \textbf{Listening} \\
        \midrule
        \textbf{Idle} &
        Agent: ``Haven't you heard? The board is cracking down on noise complaints.'' &
        User: ``But if I do, I should ask your help with setting it up.''\newline
        User: ``You know what I mean?'' \\
        \addlinespace
        \textbf{Nod} &
        User: ``The original was just so good.''\newline
        \textbf{Agent: ``Yeah, definitely.''} &
        Agent: ``That must have been exciting.''\newline
        \textbf{User: ``Everyone is so happy to see each other and catch up.''} \\
        \addlinespace
        \textbf{Shake} &
        \textbf{Agent: ``No, no, no. It's probably just a tricky language.''} &
        User: ``But she also told me to never ignore official advice.''\newline
        \textbf{User: ``That's almost impossible to do right.''} \\
        \addlinespace
        \textbf{Tilt} &
        User: ``What should I do?''\newline
        \textbf{Agent: ``Hmm, I'm not sure about that. Maybe they were just having a tough day.''} &
        User: ``And I waited there for forty-five minutes!''\newline
        \textbf{User: ``Then a girl walked by and said she saw him laughing with someone else in a coffee shop.''} \\
        \bottomrule
    \end{tabularx}
    }
    \caption{SemReact dataset examples and their corresponding dialogue excerpts.
    Top: agent facial-motion sequences for idle, nod, shake, and tilt behaviors during speaking (left) and listening (right).
    Bottom: dialogue excerpts for each behavior and conversational role.
    Bold utterances highlight the agent's speech accompanying a speaking gesture or the user's speech associated with a listening reaction.}
    \label{fig:semreact_examples}
\end{figure}

SemReact combines augmented real motion with synthetic semantic control.
The resulting dialogue--reaction--motion alignment supervises behavior selection and continuous motion realization.
Figure~\ref{fig:semreact_pipeline} summarizes how generated dialogue and synthesized speech are paired with semantically appropriate reaction motion.
We first describe the real-motion bank, followed by dialogue generation, speech synthesis, motion composition, feature extraction, and data packing for the synthetic component.
The synthetic corpus contains 18,075 accepted dialogue scripts; Table~\ref{tab:semreact_splits} summarizes its training, validation, and test partitions.
Figure~\ref{fig:semreact_examples} presents SemReact motion examples together with the corresponding dialogue excerpts for speaking and listening.

\subsection{CCDb-HG Real-Motion Bank}
\label{app:semreact_ccdb}

\para{Data Source.}
Our CCDb-HG~\citep{vuillecard2024ccdb} processing covers 55 dyadic conversations involving 22 participants, recorded with two synchronized cameras.
Training motion augmentation and CCDb-derived reference motion clips draw only from the training bank.

\para{Motion Extraction.}
We fit SMIRK~\citep{retsinas20243d} to each full video, express head pose relative to the session mean, apply a 6\,Hz low-pass filter, and resample the trajectories to 25\,fps.
The ELAN head-nodding, head-shake, and head-tilt tiers provide gesture labels, while utterance tiers provide the own-speech flag.
These trajectories supply observed gesture realizations for motion augmentation; the synthetic dialogue procedure below supplies explicit context--reaction associations.

\subsection{Dialogue Generation and Reaction Verification}
\label{app:semreact_dialogues}

\para{Reaction-Conditioned Dialogue Generation.}
We first sample a reaction budget for each dialogue: two or three listening nods, one or two listening shakes, one or two listening tilts, and one or two speaking nods or shakes.
A separate budget requests zero or one speaking tilt, for utterances expressing uncertainty, hedging, or thinking aloud.
Qwen2.5-14B-Instruct then generates exchanges designed to motivate these reactions, using a sampling temperature of 0.95, top-$p$ of 0.95, and a maximum output length of 1,800 tokens.
We vary the generation prompts over 150 topics, 14 tones, 15 registers, 10 relationship types, and six stances.
The requested dialogue length is three, five, or seven turns, sampled with probabilities 0.30, 0.45, and 0.25, respectively.
Each sentence carries its text, reaction class, and intensity.
A reaction attached to the user's sentence specifies the avatar's listener response; one attached to the avatar's sentence specifies a gesture during its own speech.
The reaction vocabulary is idle, nod, shake, and tilt, with the internal \texttt{none} label mapped to idle.
All four reaction classes are available in both speaking and listening states.

Scripts must have a valid structured format, three to nine alternating turns beginning with the user, one to six sentences per turn, and 3--400 characters per sentence.
At least one user sentence must carry a non-idle reaction.

\para{Separate Reaction Verification.}
The same language model performs a separate verification pass at temperature zero, with a maximum output length of four tokens.
It predicts the reaction from the current sentence and the immediately preceding sentence, without receiving the proposed tag.
When the two predictions disagree, we demote the tag to idle rather than substitute a different gesture.
Appendix~\ref{app:semreact_prompts} summarizes the generation and role-specific verification tasks, prompt slots, and output-validation rules.

\subsection{Speech Synthesis and Temporal Alignment}
\label{app:semreact_audio}

Kokoro-82M synthesizes the two speakers into separate 24\,kHz mono tracks on a shared timeline.
The tracks alternate: while one speaker talks, the other track contains silence.
Thus, although the representation contains both sides of the conversation, the synthesized corpus does not include overlapping speech.
We use 28 English voice packs.
With probability 0.7, a speaker's voice is a blend of two or three packs with weights sampled from a symmetric Dirichlet distribution with concentration 1.5.
Speech rate is sampled independently for each speaker from $\mathcal{U}(0.9,1.1)$.

The initial silence is sampled from $\mathcal{U}(0.2,0.5)$\,s.
Inter-sentence pauses are sampled from $\mathcal{U}(0.45,0.9)$\,s after questions and $\mathcal{U}(0.2,0.6)$\,s otherwise; turn transitions use $\mathcal{U}(0.18,0.55)$\,s, followed by a final silence of $\mathcal{U}(0.4,0.9)$\,s.
Clips shorter than 6\,s are rejected, and clips longer than 45\,s are truncated at a turn boundary.
Audio is peak-normalized to 0.95.
We generate one audio realization per dialogue, yielding 36,150 waveform files, and apply signal augmentation during data loading rather than baking it into the stored waveforms.

Sentence timestamps define reaction anchors and the avatar's own-speech flag at 25\,fps.
The speaking intervals are expanded by one frame before onset and two frames after offset to accommodate transitions.
Throughout the pipeline, the avatar's audio is the self stream and the user's audio is the partner stream.

\subsection{Continuous Motion and Gesture Composition}
\label{app:semreact_motion}

\para{Audio-Driven Base Motion.}
UniLS~\citep{chu2026unils} processes both audio tracks to generate a continuous base motion sequence for the entire dialogue.
Generating the session in one pass avoids splicing independently generated turns.
At 25\,fps, each frame contains 108 parameters: 100 expression coefficients, three head-rotation parameters, one jaw-opening parameter, and four eye-gaze parameters.
The sequence initially contains $\lceil25d\rceil$ frames for dialogue duration $d$ in seconds.

\para{Reaction Primitives and Placement.}
Table~\ref{tab:semreact_primitives} summarizes the analytic primitives added to the base sequence.
The reaction layer adds labeled phasic gestures; separate idle and gaze layers are disabled because the base already supplies continuous background motion.
Per-dialogue persona parameters vary expressivity, tempo, blink rate, and reaction latency.
Listener reactions begin after the corresponding sentence ends, with delay
\begin{equation}
    \delta = \max(0.06,\mu+0.1\epsilon)\ \mathrm{s},
    \qquad \mu\sim\mathcal{U}(0.18,0.38),\quad
    \epsilon\sim\mathcal{N}(0,1).
    \label{eq:semreact_latency}
\end{equation}
Reactions are spaced at least 1.2\,s apart, and each verified trigger is retained by the compositor.
Speaking gestures use the same primitives, with onset anchored at 25\% of the avatar's sentence duration.
Head rotations are clipped to $[-0.35,0.50]$\,rad in pitch, $\pm0.60$\,rad in yaw, and $\pm0.40$\,rad in roll; jaw opening is clipped to $[0,0.40]$\,rad.

\begin{table}[t]
    \centering
    \small
    \caption{Synthetic reaction primitives. Here $a$ is the sentence-level intensity parameter and $\mathcal{U}$ denotes uniform sampling. Frequencies for nods are additionally modulated by the sampled persona tempo.}
    \label{tab:semreact_primitives}
    \begin{tabular}{@{}p{0.08\linewidth}p{0.22\linewidth}p{0.57\linewidth}@{}}
        \hline
        Reaction & Motion profile & Parameters \\
        \hline
        Nod & \raggedright Downward raised-cosine pitch cycles & 2--3 cycles; $\mathcal{U}(1.7,2.5)$\,Hz; amplitude $0.11+0.025a$\,rad; a long blink with probability 0.5. \\
        Shake & \raggedright Damped yaw oscillation & 2--4 cycles; $\mathcal{U}(1.6,2.8)$\,Hz; amplitude $0.13+0.05a$\,rad; decay parameter 0.4 with a log-declination envelope. \\
        Tilt & \raggedright Roll movement, hold, and return & Positive roll; amplitude $0.08+0.05a$\,rad; rise $\mathcal{U}(0.5,0.9)$\,s and return $\mathcal{U}(0.6,1.0)$\,s. \\
        \hline
    \end{tabular}
\end{table}

\para{Blending and Motion Variants.}
To keep listener gestures distinct from background head sway, we scale the base head motion to 60\% during listening, with a six-frame cosine transition between speaking and listening states.
Let $\mathbf{r}^{\mathrm{base}}_f$ be the base head rotation, $\Delta\mathbf{r}_f$ the reaction offset, and $\tilde{s}_f\in[0,1]$ the smoothed own-speech flag at frame $f$.
The composed head motion is
\begin{equation}
    \mathbf{r}_f = \left[1-0.4(1-\tilde{s}_f)\right]
    \mathbf{r}^{\mathrm{base}}_f+\Delta\mathbf{r}_f.
    \label{eq:semreact_blend}
\end{equation}
The base head motion retains its full scale during speech.
We sample two gesture realizations for each dialogue while sharing the same audio and conversational hidden states.

\subsection{Conversational Features and Packed Representation}
\label{app:semreact_features}

We encode both audio tracks with Mimi and extract 4,096-dimensional temporal hidden states from frozen PersonaPlex-7B at 12.5\,Hz.
The avatar's codes occupy the self stream and the user's codes occupy the partner stream.
Feature extraction runs in evaluation mode without gradients, and the resulting hidden states are stored in bfloat16.
The conversational model is not executed or fine-tuned during motion-model training.

For a hidden-state sequence of length $\SequenceLength$, motion and frame-level labels are truncated or padded by repeating the final frame to obtain exactly $2\SequenceLength$ frames.
Using zero-based target positions, hidden state $\DialogueHidden{p}$ is aligned with motion frames $2p$ and $2p+1$, corresponding to 80\,ms.
This is an output-position index: the first motion pair in the buffer at processing step $\StreamIndex$ has target position $p=\StreamIndex+1$.
Each packed session contains hidden states of shape $\SequenceLength\times4096$, motion of shape $2\SequenceLength\times108$, and vectors of reaction labels and own-speech flags, each of length $2\SequenceLength$.
Motion is packed in float32 and both label vectors in unsigned 8-bit format.
Sessions are stored in shards of up to 128 items.

\subsection{Data Partitions and Reproducibility}
\label{app:semreact_splits}

We assign 17,388 dialogues to training, 299 to validation, and 388 to testing, as summarized in Table~\ref{tab:semreact_splits}.
All motion realizations of a dialogue remain in the same partition, preventing its dialogue content, audio, and hidden states from crossing partitions.
The validation set is used for checkpoint selection.
Evaluation uses one motion realization per test dialogue, avoiding duplicate evaluation of the same audio and dialogue with two gesture variants.

\begin{table}[t]
    \centering
    \caption{Synthetic corpus partitions. All motion realizations of a dialogue share one partition.}
    \label{tab:semreact_splits}
    \begin{tabular}{lr}
        \hline
        Split & Dialogues \\
        \hline
        Training & 17,388 \\
        Validation & 299 \\
        Test & 388 \\
        \hline
        Total & 18,075 \\
        \hline
    \end{tabular}
\end{table}

The synthesis configuration uses one audio realization and two motion realizations per dialogue, with signal augmentation disabled during waveform creation.
Motion random seeds are derived from a cyclic redundancy check of the session identifier.
Audio seeds use Python's hash function, so reproducing identical waveforms also requires fixing \texttt{PYTHONHASHSEED}.

\subsection{Style Reference Bank Construction}
\label{app:semreact_style_references}

The style reference bank is constructed from reference motion clips drawn from the CCDb-HG training bank (Appendix~\ref{app:semreact_ccdb}).
It covers eight state--reaction combinations: idle, nod, shake, and tilt for both speaking and listening.
Each reference clip comprises 100 frames (4\,s) of 108-dimensional motion and is indexed by its speaking/listening state and reaction class.
All references come from the training partition.

The generator applies its shared framewise linear motion projection $\MotionProjection$ to the selected clip, yielding 100 style tokens of width 512.
The same projection is used for persona, history, and noisy motion frames; there is no separate style encoder.
Separate speaking and listening references allow the same reaction class to condition motion appropriate to the avatar's current interaction state.

\section{Semantic Context-Guided Non-Verbal Behavior: Implementation Details}
\label{app:semantic_behavior}

This section provides implementation details for Section~\ref{sec:semantic_behavior}; joint training is described in Appendix~\ref{app:motion_training}.
The generator receives the hidden-state window separately from $\MotionConditions$, which comprises history tokens, fixed persona tokens, routed style tokens, and framewise behavior controls.

\subsection{Semantic Behavior Router: State and Reaction Decisions}
\label{app:semantic_behavior_router}

The semantic behavior router determines the avatar's speaking/listening state $\DialogueState{\StreamIndex}\in\{0,1\}$ and reaction from recent conversational hidden states.
Its state decision and accepted reaction form the pair $(\DialogueState{\StreamIndex},\AcceptedBehavior{\StreamIndex})$ used for state-specific style selection.
The final pair receives four expression--jaw updates in both speaking and listening states (Appendix~\ref{app:motion_updates}).
Equation~\ref{eq:semantic_router} defines the two probability outputs.

At processing step $\StreamIndex$, the causal memory $\ContextMemory{\StreamIndex}=[\DialogueHidden{\StreamIndex-\ContextLength+2},\ldots,\DialogueHidden{\StreamIndex+1}]$ contains the latest $\ContextLength=25$ conversational states computed from received user audio tokens at 80\,ms intervals.
Its newest state reads input row $\StreamIndex$ and does not depend on predicted user audio tokens at future positions.
Only the router reads this longer history; the motion generator retains its short run-ahead speech window.
The router shares the generator's speech projection and applies layer normalization followed by an MLP with separate reaction and speaking/listening output heads to the pooled context (Eq.~\ref{eq:semantic_router}).
At inference, a proposed class change must persist for two consecutive streaming steps before it becomes the accepted reaction $\AcceptedBehavior{\StreamIndex}$.

\subsection{Style Token Routing and Framewise Controls}
\label{app:semantic_behavior_control}

The bank lookup $\StyleRouter$ selects the reference motion clip matching the accepted reaction and the router's speaking/listening decision (Appendix~\ref{app:semreact_style_references}).
The shared linear projection $\MotionProjection$ maps its 100 frames to 100 style tokens, as defined in Eq.~\ref{eq:style_prompt_routing}.
These tokens are concatenated with 100 session-fixed persona tokens for motion conditioning.
When the selected reference changes, the previous style-token sequence is cached as $\PreviousStyleTokens{\StreamIndex}$ and held as the source of the transition.
The weight $\FadeWeight{\StreamIndex}\in[0,1]$ starts at zero and increases by $1/4$ per streaming step, capped at one, giving a four-step linear crossfade:
\begin{equation}
    \BlendedStyleTokens{\StreamIndex}
    = \FadeWeight{\StreamIndex}\StyleTokens{\StreamIndex}
    +(1-\FadeWeight{\StreamIndex})\PreviousStyleTokens{\StreamIndex}.
    \label{eq:style_token_crossfade}
\end{equation}
After the transition, $\BlendedStyleTokens{\StreamIndex}=\StyleTokens{\StreamIndex}$; persona tokens remain fixed throughout.
For example, speaking and listening nods use distinct style references with the same persona tokens.

\subsection{Reaction Progression and Temporal Continuity}
\label{app:semantic_behavior_progression}

To represent a reaction as an evolving movement, the generator receives its class and elapsed phase at each motion frame.
At motion-frame index $\MotionFrameIndex$, the token $\MotionToken{\MotionFrameIndex}\in\R^{\MotionWidth}$ combines projected noisy motion with token-type and flow-time embeddings.
It is distinct from the raw flow-space motion $\NoisyMotion_{\MotionFrameIndex}\in\R^{\MotionDim}$.
Let $\AcceptedBehavior{\MotionFrameIndex}$ and $\DialogueState{\MotionFrameIndex}$ denote the behavior and speaking/listening state aligned to that frame, and let $\GesturePhase{\MotionFrameIndex}$ count frames since gesture onset, capped at 49.
We add the corresponding learned embeddings:
\begin{equation}
    \MotionToken{\MotionFrameIndex} \leftarrow \MotionToken{\MotionFrameIndex}
    + \BehaviorEmbedding(\AcceptedBehavior{\MotionFrameIndex})
    + \StateEmbedding(\DialogueState{\MotionFrameIndex})
    + \PhaseEmbedding(\GesturePhase{\MotionFrameIndex}),
    \label{eq:behavior_phase_conditioning}
\end{equation}
where $\BehaviorEmbedding$, $\StateEmbedding$, and $\PhaseEmbedding$ map behavior, the speaking/listening state, and gesture phase to $\MotionWidth$-dimensional vectors, respectively.
The phase counter resets on an accepted behavior change, remains zero for idle, and otherwise advances by two frames per streaming tick up to the cap.
Within each emitted pair, the second frame advances the phase by one, also capped at 49.
An unchanged non-idle prediction therefore continues the existing reaction rather than resetting its onset at every tick.
These controls allow the generator to distinguish gesture onset, progression, and return from the training trajectories.
When completed frames enter the motion history, their labels carry over to the history tokens and remain part of the motion conditioning $\MotionConditions$.
Consecutive-decision confirmation suppresses transient class changes, phase conditioning preserves reaction progress, and the style-token crossfade smooths accepted reference changes.
Together, these mechanisms support coherent reaction execution across rolling chunks.
They do not impose a fixed minimum gesture duration or block a sustained new decision until the current gesture finishes: an accepted class change resets the phase, and reaching the phase cap does not itself trigger a return to idle.

\section{SemReact Prompt Specifications}
\label{app:semreact_prompts}

This section summarizes the generation and verification tasks, their input fields, and output formats for the synthetic dialogue component described in Appendix~\ref{app:semreact_dialogues}.
Both stages use Qwen2.5-14B-Instruct served with vLLM.
The reaction vocabulary is idle, nod, shake, and tilt: nod expresses agreement or acknowledgement, shake expresses negation or disagreement, and tilt expresses uncertainty or deliberation.
The internal \texttt{none} label represents idle.

\subsection{Dialogue Generation}
\label{app:semreact_generation_prompts}

Generation uses temperature 0.95, top-$p$ 0.95, and a maximum output length of 1,800 tokens.
The instructions request a natural spoken-English exchange between USER and AGENT, with alternating turns beginning with USER and a sampled conversational context.
A reaction budget specifies how many user sentences should motivate each listener reaction and how many agent sentences should accompany speaking gestures.
User-sentence tags describe the agent's listener reaction; agent-sentence tags describe its own speaking gesture.
Each sentence contains \texttt{text}, \texttt{tag}, and \texttt{intensity}, organized within a JSON \texttt{turns} array.
The requested intensity range is one to three, with intensity three used at most once.
Prompt slots and structural acceptance criteria are detailed below.

\subsection{Prompt Slots and Output Validation}
\label{app:semreact_prompt_slots}

\para{Conversational Context.}
Relationship, setting, topic, and user tone are sampled independently and uniformly from their respective candidate lists.
The ten relationship values are a close friend, a coworker, a sibling, a roommate, a neighbor, an old classmate, a partner, a cousin, a gym buddy, and a fellow parent.
The 14 tone values are excited, worried, amused, frustrated, nostalgic, proud, embarrassed, sad, relieved, angry, matter-of-fact, confused, hopeful, and sarcastic.
The 15 settings include casual conversation, catching up, sharing news, telling a story, asking for advice, and describing future plans.
The 150 topics cover everyday experiences such as work stress, restaurants, childhood memories, moving city, cooking failures, travel, relationships, and pets.
The setting and topic examples here illustrate their coverage rather than enumerate the full candidate lists.

\para{Avatar Stance.}
Table~\ref{tab:semreact_prompt_stances} gives the six stance values, their probabilities, and the associated instruction inserted into the generation prompt.
The stance-specific terms describe desired conversational situations; the sentence-level output tags remain \texttt{nod}, \texttt{shake}, \texttt{tilt}, and \texttt{none}.

\begin{table}[t]
    \centering
    \small
    \caption{Avatar stance slots for dialogue generation. Each row provides the values of \texttt{\{stance\}} and \texttt{\{stance\_req\}}, together with its sampling probability.}
    \label{tab:semreact_prompt_stances}
    \begin{tabular}{@{}p{0.25\linewidth}rp{0.57\linewidth}@{}}
        \hline
        Stance & Prob. & Required conversational situation \\
        \hline
        \raggedright supportive and agreeing & 0.24 & include at least one \texttt{agree\_strong} moment \\
        \raggedright skeptical & 0.18 & include at least one disagree moment (the speaker says something the listener clearly would not accept) \\
        \raggedright astonished & 0.18 & include at least one disbelief moment (the speaker reveals something hard to believe) \\
        \raggedright empathetic & 0.16 & include at least one sympathy moment \\
        \raggedright being asked for their opinion & 0.14 & include at least one think moment (the speaker asks the listener something rhetorical or direct) \\
        \raggedright engaged but sometimes lost & 0.10 & include at least one confusion moment \\
        \hline
    \end{tabular}
\end{table}

\para{Length and Reaction Budgets.}
We sample three, five, or seven turns with probabilities 0.30, 0.45, and 0.25.
Their respective hint strings are ``short exchange'', ``normal back-and-forth'', and ``longer conversation''.
Each reaction-budget slot is sampled independently with equal probability between its two values: listening nods in $\{2,3\}$, listening shakes in $\{1,2\}$, listening tilts in $\{1,2\}$, and speaking gestures in $\{1,2\}$.
The additional speaking-tilt budget is zero or one with probabilities 0.4 and 0.6, respectively.
The prompt requests intensity values from one to three, with at most one sentence assigned intensity three.

\para{Structured Output and Acceptance.}
The model returns a JSON object containing a \texttt{turns} array.
Each turn has a \texttt{speaker} and a \texttt{sentences} array; each sentence contains \texttt{text}, \texttt{tag}, and \texttt{intensity}.
The parser accepts three to nine alternating turns beginning with the user, one to six sentences per turn, and text lengths of 3--400 characters.
It additionally requires at least two user sentences and at least one non-idle user-sentence tag, and clips intensity to $[1,3]$.
These structural checks are followed by the semantic verification below.

\subsection{Reaction Verification}
\label{app:semreact_verification_prompts}

Verification uses temperature zero and a maximum output length of four tokens.
Each non-idle proposal is checked using the current sentence and, when available, the immediately preceding sentence.
The proposed tag is withheld from the verifier.
For user sentences, the task is to infer the agent's listener reaction; for agent sentences, it is to infer the speaker's accompanying gesture.
The verifier returns one class from \texttt{nod}, \texttt{shake}, \texttt{tilt}, or \texttt{none}.
We parse the response using \verb+\b(nod|shake|tilt|none)\b+.
A proposal is retained only when the parsed class agrees with its generation tag; disagreement maps it to idle rather than another non-idle gesture.

\section{Streaming Generation Details}
\label{app:streaming_details}

\subsection{Motion Generator}
\label{app:motion_architecture}

\begin{figure}[htbp]
    \centering
    \includegraphics[width=\linewidth]{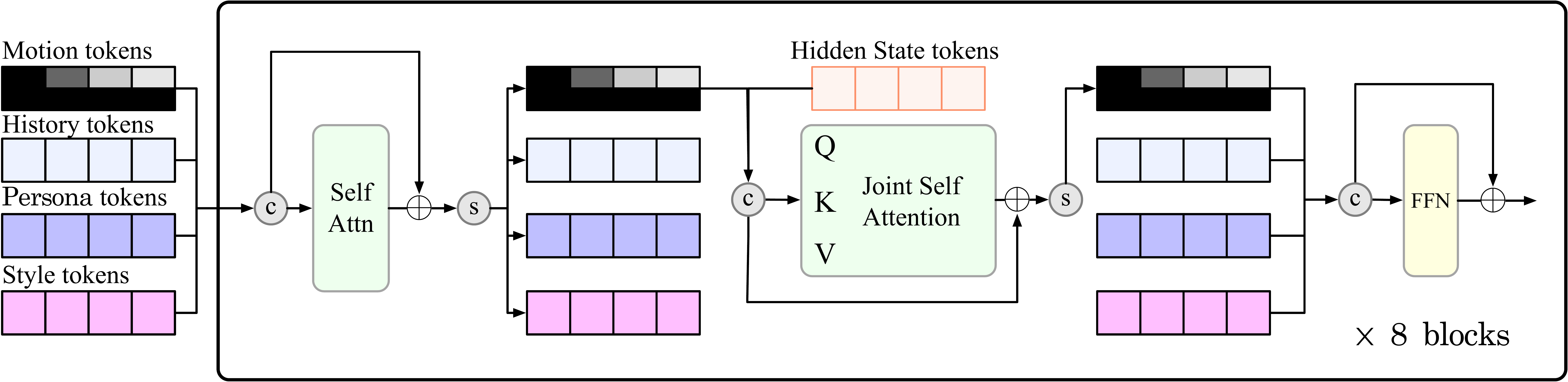}
    \caption{Motion generator architecture.
    Self-attention combines motion, history, persona, and style tokens, while joint self-attention incorporates hidden states into the motion tokens.
    The block is repeated eight times.
    Symbols $c$ and $s$ denote concatenation and splitting; token counts are schematic.}
    \label{fig:motion_generator_architecture}
\end{figure}

Figure~\ref{fig:motion_generator_architecture} illustrates the motion generator and its conditioning inputs.
The shared motion generator $\VelocityField$, parameterized by trainable weights $\MotionParams$, has eight transformer layers, width $\MotionWidth=512$, and eight attention heads.
The generator maintains $\ActiveFrameCount=8$ noisy motion frames, grouped into $\QueueLength=4$ pairs.
Each pair contains two tokens and corresponds to one 80\,ms streaming step.
Persona tokens $\PersonaTokens$ and style tokens $\StyleTokens{\StreamIndex}$ each represent a 100-frame reference motion clip, one token per frame.
The persona reference and its tokens remain fixed throughout a session.
The bank lookup $\StyleRouter$ selects the style reference matching the reaction class and the router's speaking/listening decision.
History tokens represent the eight previous completed frames, while eight motion tokens represent the current noisy frames.
All four token types use the same framewise linear projection $\MotionProjection$ from $\MotionDim=108$ coordinates to width $\MotionWidth=512$, implemented by a single shared \texttt{nn.Linear(108, 512)} layer.
Token-type embeddings distinguish motion, history, and reference inputs; additional reference-type embeddings distinguish persona tokens from style tokens.
These are continuous feature vectors, each representing one 108-dimensional FLAME frame.
Each motion token $\MotionToken{\MotionFrameIndex}$ combines projected noisy coordinates with token-type and group-specific flow-time embeddings.
Embeddings of behavior, the speaking/listening state, and gesture phase provide the controls described in Section~\ref{sec:semantic_behavior}.

Within each layer, motion self-attention combines motion tokens with history, persona, and style tokens.
Joint self-attention operates on the motion tokens and the four hidden states projected by $\SpeechProjection\in\R^{\MotionWidth\times\SpeechDim}$.
We denote the history tokens, persona tokens, style tokens, and framewise controls collectively by $\MotionConditions$, with the hidden-state window supplied separately.
Only the motion outputs are retained as residual updates, followed by a feed-forward block.
The longer conversational memory is used by the router; its states are not added to this short speech window.

\subsection{Run-Ahead Buffer and Cache}
\label{app:run_ahead_details}

Figure~\ref{fig:run_ahead_buffer_cache} illustrates how sequential warm-up prepares the token buffer and how batched run-ahead $\BatchedRunAhead$ recomputes the hidden states after receiving new user audio tokens.
During warm-up, each hidden state computed by the temporal transformer $\TemporalTransformer$ is passed to the depth transformer $\DepthTransformer$, and the resulting predicted user audio tokens and avatar audio tokens become inputs for the next step.
This sequential rollout continues until the buffer contains the required input window.
The buffer stores avatar audio and text tokens, predicted user audio tokens, and incomplete entries for delayed codebooks.
At each step, the predicted user audio tokens for the current step are replaced by the received user audio tokens as described in Section~\ref{sec:run_ahead}, while the avatar audio and text tokens remain fixed.
If codebook entries required for the next window are incomplete, they are filled sequentially before batched evaluation.

\begin{figure}[t]
    \centering
    \includegraphics[width=\linewidth]{Figures/Method1.pdf}
    \caption{Run-ahead prediction and hidden-state recomputation.
    \textbf{(a)} Sequential self-replay prepares the token buffer during warm-up.
    \textbf{(b)} Received user audio tokens replace their predictions while buffered avatar tokens remain fixed.
    The temporal transformer $\TemporalTransformer$ recomputes the hidden-state window in one causal pass; only its last state is passed to the depth transformer $\DepthTransformer$ (depformer) to extend the buffer.
    In $\DialogueHidden{i\mid\StreamIndex}$, $i$ denotes the output position and $\StreamIndex$ the processing step at which the state is computed.
    Cyan tokens denote predicted user audio, red-bordered tokens received user audio, and black-bordered tokens avatar audio.
    Earlier history is reused through the KV cache, while the window's cache entries are recomputed.}
    \label{fig:run_ahead_buffer_cache}
\end{figure}

Once the buffer is filled, the temporal transformer $\TemporalTransformer$ can read all rows in the window together: their token values are already known, and causal attention preserves their temporal order.
At processing step $\StreamIndex$, a single causal transformer pass reads input rows $\StreamIndex,\ldots,\StreamIndex+\RunAheadHorizon$ from $\UpdatedTokenBuffer_{\StreamIndex}$ using $\SpeechCache$ and produces states at positions $\StreamIndex+1,\ldots,\StreamIndex+\RunAheadHorizon+1$.
Only the last state $\ProvisionalHidden_{\StreamIndex+\RunAheadHorizon+1\mid\StreamIndex}$ is passed to the output heads and depth transformer $\DepthTransformer$ to generate $\PredictedUserAudioTokens{\StreamIndex+\RunAheadHorizon+1}$ and $\AvatarTokens{\StreamIndex+\RunAheadHorizon+1}$, as returned by $\BatchedRunAhead$ in Section~\ref{sec:run_ahead}.
Here $\AvatarTokens{\StreamIndex+\RunAheadHorizon+1}$ denotes avatar audio tokens; the accompanying text tokens are also generated but omitted from the notation.
Appending the generated audio and text tokens extends the buffer to position $\StreamIndex+\RunAheadHorizon+1$.
For the same tokens, cache, positions, and attention mask, this computation matches sequential evaluation up to numerical precision.
This equivalence concerns evaluation of the same buffered inputs; it does not imply equivalence to decoding after receiving future user audio tokens.
Appendix~\ref{app:latency} reports timing measurements with explicit processing boundaries.

The KV cache of $\TemporalTransformer$ stores each layer's keys and values for earlier input rows, allowing later rows to attend to that history without recomputing it.
Entries preceding $\StreamIndex$ depend on received user audio tokens and committed avatar tokens, so they can be reused.
After replacing $\PredictedUserAudioTokens{\StreamIndex}$ with $\UserAudioTokens{\StreamIndex}$, the transformer recomputes the keys and values at input position $\StreamIndex$.
Through causal attention, this update can affect hidden states and keys and values at all later positions in the window, so the window is recomputed while $\SpeechCache$ is reused.
Cache entries correspond to input positions, whereas hidden-state indices denote next-token positions.
After extracting the hidden-state window, we retain the recomputed keys and values for input position $\StreamIndex$ in $\SpeechCache$, extending the history of received inputs for the next pass.
At step $\StreamIndex+1$, we use the received user audio tokens to recompute $\DialogueHidden{\StreamIndex+2}$ and the overlapping future hidden states:
\begin{equation}
    \begin{aligned}
        \ProvisionalHidden_{\StreamIndex+2\mid\StreamIndex}&\rightarrow\DialogueHidden{\StreamIndex+2}, \\
        \ProvisionalHidden_{\StreamIndex+\FutureOffset\mid\StreamIndex}
        &\rightarrow\ProvisionalHidden_{\StreamIndex+\FutureOffset\mid\StreamIndex+1},
        \quad \FutureOffset=3,\ldots,\RunAheadHorizon+1.
    \end{aligned}
    \label{eq:run_ahead_update}
\end{equation}
The new window also adds $\ProvisionalHidden_{\StreamIndex+\RunAheadHorizon+2\mid\StreamIndex+1}$ at its far end.
Only the cache entry for the input row with newly received user audio tokens is retained as completed history; future entries remain provisional and are overwritten on the next pass.
Because the cache is a ring buffer, a future-position write can land on a slot that still holds valid history.
We snapshot such slots before the pass and restore them afterwards.
The token buffer and motion buffer persist across steps.

\subsection{Consistency Between Committed Speech and Recomputed States}
\label{app:run_ahead_consistency}

As described in Section~\ref{sec:run_ahead}, the committed avatar tokens remain fixed while received user audio changes the hidden-state window $\OnlineSpeechWindow_{\StreamIndex}$.
We first compare the distributions underlying actually generated and committed audio before and after hidden-state recomputation.
We also report counterfactual speech comparisons.

\para{Distribution Consistency of Committed Audio.}
We evaluate run-ahead horizons $\RunAheadHorizon=0,\ldots,6$ on 22 independent Seamless Interaction parent sessions, after removing duplicate parents.
For each horizon, we retain the avatar audio tokens actually generated and committed by the model and compare their conditional distributions at commitment and after hidden-state recomputation using received user audio.
The committed tokens themselves remain unchanged throughout this comparison.
Semantic results use audio codebook 0, while acoustic results average codebooks 1--7.

For distributions $p$ before recomputation and $q$ after recomputation, we report Jensen--Shannon (JS) divergence in bits, overlap $\sum_v\min\{p(v),q(v)\}=1-\operatorname{TV}(p,q)$, and top-1 agreement between their most likely tokens.
Here $\operatorname{TV}$ denotes total variation distance.
For the same committed token $a$, we define the change in negative log-likelihood as $\Delta\mathrm{NLL}=\mathrm{NLL}_{\mathrm{updated}}-\mathrm{NLL}_{\mathrm{commit}}=\log p(a)-\log q(a)$, using natural logarithms (nat/token).
A positive value means that the updated distribution assigns a lower probability to the committed token.
Top-1 agreement measures agreement between these distributions' predictions, rather than changes to the committed audio.

Table~\ref{tab:committed_audio_distribution} aggregates the full evaluated sequences, with 95\% confidence intervals obtained by bootstrapping sessions.
At $\RunAheadHorizon=0$, the comparison contains 2,816 semantic and 19,712 acoustic token positions; each nonzero horizon uses 2,794 and 19,558, respectively.
The zero-horizon self-comparison yields JS divergence at numerical precision.
At the default $\RunAheadHorizon=3$, semantic and acoustic overlap are 0.9666 and 0.9353, with JS divergences of 0.005987 and 0.017798 bits.
These results support approximate distributional consistency after recomputation, with larger changes in acoustic than semantic distributions.
\begin{table}[!htbp]
    \centering
    \small
    \caption{Distribution consistency of committed avatar audio before and after hidden-state recomputation. Brackets show 95\% confidence intervals from session-level bootstrap; gray marks the default horizon.}
    \label{tab:committed_audio_distribution}
    \footnotesize
    \setlength{\tabcolsep}{3pt}
    \renewcommand{\arraystretch}{1.05}
    \begin{tabular}{@{}clcccc@{}}
        \toprule
        $\RunAheadHorizon$ & Codebook & JS (bits) $\downarrow$ & Overlap $\uparrow$ & Top-1 (\%) $\uparrow$ & $\Delta$NLL (nat/token) \\
        \midrule
        0 & Semantic & $0\,[0,0]$ & 1.000000 & 100.000 & $0.000000\,[0,0]$ \\
         & Acoustic & $0\,[0,0]$ & 1.000000 & 100.000 & $0.000000\,[0,0]$ \\
        \midrule
        1 & Semantic & $0\,[0,0]$ & 1.000000 & 100.000 & $0.000000\,[0,0]$ \\
         & Acoustic & $0\,[0,0]$ & 1.000000 & 100.000 & $0.000000\,[0,0]$ \\
        \midrule
        2 & Semantic & $0.004508\,[0.002887,0.006455]$ & 0.972877 & 98.210 & $0.010077\,[0.001076,0.022214]$ \\
         & Acoustic & $0.014498\,[0.010642,0.018533]$ & 0.947142 & 92.402 & $0.043631\,[0.032285,0.055479]$ \\
        \midrule
        \rowcolor{lightgray!40}
        3 & Semantic & $0.005987\,[0.004101,0.007986]$ & 0.966600 & 96.886 & $0.014805\,[0.006888,0.023748]$ \\
        \rowcolor{lightgray!40}
         & Acoustic & $0.017798\,[0.013137,0.022747]$ & 0.935278 & 90.526 & $0.058402\,[0.041641,0.076760]$ \\
        \midrule
        4 & Semantic & $0.004461\,[0.002933,0.006093]$ & 0.973623 & 97.387 & $0.009967\,[0.006056,0.014252]$ \\
         & Acoustic & $0.012939\,[0.009751,0.016499]$ & 0.951011 & 93.092 & $0.040853\,[0.029861,0.053107]$ \\
        \midrule
        5 & Semantic & $0.008316\,[0.005685,0.011200]$ & 0.960032 & 96.349 & $0.025582\,[0.017008,0.036535]$ \\
         & Acoustic & $0.017673\,[0.013431,0.022039]$ & 0.935669 & 90.464 & $0.056623\,[0.043009,0.070145]$ \\
        \midrule
        6 & Semantic & $0.006963\,[0.004897,0.009307]$ & 0.963124 & 96.206 & $0.024076\,[0.014488,0.033951]$ \\
         & Acoustic & $0.017267\,[0.011827,0.023145]$ & 0.933940 & 89.866 & $0.059146\,[0.038516,0.082458]$ \\
        \bottomrule
    \end{tabular}
    \par\vspace{2pt}
    Values below $10^{-16}$ are shown as zero.
\end{table}


\para{Greedy Speech Agreement.}
We evaluate 50 Seamless Interaction validation windows containing 7,550 frames.
This separate counterfactual comparison allows the avatar predictions to differ; its disagreement rates do not measure audio--motion mismatches.
From the same ground-truth history, branch S uses predicted user audio tokens from the speech backbone, while branch R uses received user audio tokens.
Both branches generate avatar tokens using greedy decoding.
Branch R is a counterfactual used for this evaluation; its avatar predictions do not replace committed tokens in the run-ahead system.
Table~\ref{tab:run_ahead_token_agreement} reports agreement between these branches at three future positions.
At 240\,ms, 4.3\% of semantic tokens differ, compared with 2.1\% of text tokens and 21.0\% of acoustic tokens.
Semantic-token agreement with the ground-truth conversation is nearly unchanged: 0.436 for S and 0.435 for R.
The perfect semantic agreement at 80\,ms follows from the decoder ordering, in which avatar codebooks precede user codebooks in the depth transformer.
These are greedy prediction comparisons; we did not measure token-change rates under sampling.

\begin{table}[t]
    \centering
    \small
    \caption{Counterfactual comparison of greedy avatar predictions conditioned on predicted user audio tokens (S) or received user audio tokens (R). The first three agreement columns compare these speech continuations; the last two compare semantic predictions with ground-truth tokens.}
    \label{tab:run_ahead_token_agreement}
    \begin{tabular}{cccc
        >{\centering\arraybackslash}p{0.11\linewidth}
        >{\centering\arraybackslash}p{0.11\linewidth}}
        \toprule
        & \multicolumn{3}{c}{S--R agreement} & \multicolumn{2}{c}{Semantic agreement with GT} \\
        \cmidrule(lr){2-4}\cmidrule(lr){5-6}
        Ahead (ms) & Text & Semantic & Acoustic & R & S \\
        \midrule
        +80  & 1.000 & 1.000 & 0.920 & 0.573 & 0.573 \\
        +160 & 0.993 & 0.983 & 0.839 & 0.486 & 0.487 \\
        +240 & 0.979 & 0.957 & 0.790 & 0.435 & 0.436 \\
        \bottomrule
    \end{tabular}
\end{table}

\para{Speech Distribution Overlap.}
To compare the full next-token distributions, we sample 2,080 probes from 80 clips at six-frame intervals, again starting from the same ground-truth history for each comparison.
We compare three user-input conditions against received user audio tokens (R): predicted user audio tokens (S), repetition of the tokens from the last received user audio frame (H), and tokens from an unrelated user audio frame (W).
We use the same overlap and JS metrics defined above.
Table~\ref{tab:run_ahead_distribution} also reports semantic top-1 agreement.

At 240\,ms, S retains 0.979 text overlap and 0.970 semantic overlap, with JS divergences of 0.0034 and 0.0058 bits, respectively.
For speaking frames alone, text and semantic overlap are 0.988 and 0.980 at 160\,ms, and 0.977 and 0.962 at 240\,ms.
Top-1 agreement here is measured on these 2,080 probes, separately from the greedy rollout evaluation above.
Even W retains more than 0.93 overlap at 240\,ms, suggesting that the existing dialogue history strongly constrains the immediate speech predictions.
Together, these results support short-term stability of text and semantic predictions across these input conditions, while acoustic tokens differ more between the counterfactual speech continuations.
Already committed avatar tokens cannot be revised in response to later user input.

\begin{table}[t]
    \centering
    \small
    \setlength{\tabcolsep}{5pt}
    \caption{Avatar next-token distributions relative to received user audio tokens (R). S uses predicted user audio tokens, H repeats the tokens from the last received user audio frame, and W uses tokens from an unrelated user audio frame. Overlap and semantic top-1 agreement are fractions; JS divergence is in bits.}
    \label{tab:run_ahead_distribution}
    \begin{tabular}{lcccccc}
        \toprule
        & & \multicolumn{2}{c}{Text} & \multicolumn{3}{c}{Semantic} \\
        \cmidrule(lr){3-4}\cmidrule(lr){5-7}
        User input & Ahead (ms) & Overlap & JS & Overlap & JS & Top-1 \\
        \midrule
        S: predicted & +160 & 0.990 & 0.0013 & 0.983 & 0.0026 & 0.980 \\
        S: predicted & +240 & 0.979 & 0.0034 & 0.970 & 0.0058 & 0.964 \\
        H: last received & +240 & 0.979 & 0.0032 & 0.970 & 0.0053 & 0.960 \\
        W: unrelated & +240 & 0.931 & 0.0210 & 0.934 & 0.0162 & 0.933 \\
        \bottomrule
    \end{tabular}
\end{table}

\subsection{Numerical Motion Updates}
\label{app:motion_updates}

We split each frame into expression--jaw ($\ExpressionJawGroup$, 101 coordinates) and head--eye ($\HeadEyeGroup$, seven coordinates).
The expression--jaw group includes all expression coefficients, including the upper face.
Each group has its own flow time, from noise at $\FlowTime=0$ to clean motion at $\FlowTime=1$.
At processing step $\StreamIndex$, we output the pair for position $\StreamIndex$, completed at step $\StreamIndex-1$, and use its frames to update the history tokens.
We shift the motion pairs one position forward in the buffer and add a new noise pair at the end, so the active pairs target output positions $\StreamIndex+1,\ldots,\StreamIndex+\QueueLength$.
Head--eye flow times start at $[3/4,1/2,1/4,0]$ from the first to the last pair.
The two shared updates each advance these times by $1/8$.
Expression--jaw coordinates remain at the noise source in the last three pairs and complete their denoising in the first pair.

The motion buffer contains $\QueueLength=4$ pairs, and each streaming step uses $\SharedSteps=2$ shared generator evaluations.
These evaluations update head--eye coordinates in all pairs and expression--jaw coordinates in the first pair.
Another $\ExtraSteps=2$ evaluations update only the first pair's expression--jaw coordinates in both speaking and listening states.
The expression--jaw update count is therefore
\begin{equation}
    \ExpressionJawSteps=\SharedSteps+\ExtraSteps=2+2=4,
    \label{eq:expression_jaw_refinement_budget}
\end{equation}
giving the same final-step schedule $(0,0,0,4)$ in both states.
Figure~\ref{fig:future_context_analysis}(c) illustrates this schedule.
Head--eye coordinates and flow times remain fixed during the extra evaluations.

Let $\NoisyMotion$ be the full noisy motion buffer, $\FlowTimes$ its group flow times, $\SpeechWindow$ the hidden-state window, and $\GroupStepSize$ the step size for group $\MotionGroupIndex$.
During streaming, $\SpeechWindow=\OnlineSpeechWindow_{\StreamIndex}$.
We apply Euler updates to the selected coordinates:
\begin{equation}
    \NoisyMotion_{\MotionFrameIndex}^{\MotionGroupIndex}\leftarrow
    \NoisyMotion_{\MotionFrameIndex}^{\MotionGroupIndex}
    +\GroupStepSize\left[\FrameVelocity{\MotionFrameIndex}
    (\NoisyMotion;\FlowTimes,\SpeechWindow,\MotionConditions)\right]^{\MotionGroupIndex}.
    \label{eq:group_refinement}
\end{equation}
The head--eye step size is $\HeadEyeStepSize=1/(\QueueLength\SharedSteps)=1/8$.
For expression--jaw motion, it is $\ExpressionJawStepSize=1/\ExpressionJawSteps=1/4$ in both states.
Both component groups finish at flow time one in either state.
All evaluations call the full shared generator.
Shared and extra evaluations use the same history, persona, and style tokens, current state, reaction, and phase labels, and labels carried by history tokens.
They also use the same hidden-state window.
During the extra evaluations, only the first pair's expression--jaw coordinates and their flow time advance; the head--eye coordinates and flow times remain fixed.

\subsection{Motion Training}
\label{app:motion_training}

For clean motion $\CleanMotion_{\MotionFrameIndex}$ and Gaussian noise $\MotionNoise_{\MotionFrameIndex}\sim\mathcal{N}(0,\IdentityMatrix_{\MotionDim})$, the training path is
\begin{equation}
    \NoisyMotion_{\MotionFrameIndex}^{\MotionGroupIndex}
    =(1-\FlowTime_{\MotionFrameIndex}^{\MotionGroupIndex})\MotionNoise_{\MotionFrameIndex}^{\MotionGroupIndex}
    +\FlowTime_{\MotionFrameIndex}^{\MotionGroupIndex}\CleanMotion_{\MotionFrameIndex}^{\MotionGroupIndex},
    \qquad \MotionGroupIndex\in\{\ExpressionJawGroup,\HeadEyeGroup\}.
    \label{eq:asymmetric_flow_path}
\end{equation}
The superscript $\MotionGroupIndex$ selects a group's coordinates.
Together, the groups form a noisy frame $\NoisyMotion_{\MotionFrameIndex}\in\R^{\MotionDim}$.
We train the generator to predict the displacement from noise to clean motion:
\begin{equation}
    \MotionFlowLoss=
    \E\!\left[
    \frac{1}{\ActiveFrameCount\MotionDim}\sum_{\MotionFrameIndex=1}^{\ActiveFrameCount}
    \left\|\FrameVelocity{\MotionFrameIndex}(\NoisyMotion;\FlowTimes,\SpeechWindow,\MotionConditions)
    -(\CleanMotion_{\MotionFrameIndex}-\MotionNoise_{\MotionFrameIndex})\right\|_2^2\right].
    \label{eq:rolling_fm_loss}
\end{equation}
Here $\NoisyMotion$ is the noisy motion buffer and $\FlowTimes$ collects its group flow times.
The loss averages velocity errors over all eight frames and 108 coordinates.

We sample noisy inputs directly from Eq.~\ref{eq:asymmetric_flow_path}, without unrolling the inference updates.
For head--eye motion, the four pairs start at flow times $[3/4,1/2,1/4,0]$.
We add the same random offset, uniform over $[0,1/4]$, to these times.
For expression--jaw motion, we independently sample the first pair's time uniformly over $[0,1]$ and keep the last three pairs at zero.
Both frames in a pair share their group flow times.
All coordinates receive velocity supervision, including expression--jaw coordinates at flow time zero in the last three pairs.

Each training crop contains eight history frames and two consecutive eight-frame target chunks.
We evaluate the chunks separately, using their preceding ground-truth frames to form history tokens, and average their losses.
The second chunk does not use generated motion from the first.
The discrete update counts are used only during inference.

We train the motion generator and router from scratch, keeping the speech backbone and codec frozen.
Training combines synthetic speaking motion, real conversations from Seamless Interaction, and labeled synthetic reaction motion.
Both router heads receive cross-entropy supervision through $\RoutingLoss$ in Section~\ref{sec:semantic_behavior}.

\para{Optimization.}
We randomly initialize the trainable parameters and optimize them for 100,000 steps on one H200 with a batch size of 128 crops.
We use AdamW with zero weight decay and a peak learning rate of $5\times10^{-5}$.
The learning rate increases linearly over the first 500 steps and then follows cosine decay to $5\times10^{-6}$.
We clip the global gradient norm to 1.0.
The model and data random seeds are both 1234.
Evaluation uses the non-EMA model weights saved at step 100,000.

\para{Training Data and Conditioning.}
The three training sources are synthetic speaking motion, Seamless Interaction conversations, and labeled SemReact reaction motion.
Data shards from these sources are interleaved, with sample order shuffled within each shard.
Each crop supplies eight history frames and two eight-frame target chunks; persona and style references each contain 100 frames.
The router pools 25 conversational hidden states.

\subsection{Ablation Details}
\label{app:controlled_ablation}

\para{Component References.}
The no-hidden-state-update row in Table~\ref{tab:ablation_recomputation_schedule} retains a previously recorded run on samples from the Seamless Interaction test split that keeps the first predicted hidden state for each absolute target position.
The w/o Asym. and Ours rows report the seed-s1 uniform $K=1$ evaluation and the retained main-table reference, respectively.
The uniform generator applies one refinement call at each of the four rolling positions to all coordinates, totaling four updates per pair.
This matches the expression--jaw update count of FacePlex; the head--eye schedules and total generator computation remain different.
L-FID is recomputed from pooled raw FLAME listening frames in the evaluated test-split samples.
The w/ Cross-attn and w/o SemReact rows report the corresponding conditioning and training variants; the full configuration is represented once by Ours.

\para{Rule-Based Reaction Selection.}
\phantomsection\label{app:router_rule_baseline}
Table~\ref{tab:router_rule_comparison} compares the final frozen router with a handwritten phrase-count baseline on the same 38,375 annotated reaction frames from 388 test dialogues.
Both methods are given ground-truth reaction presence and select only nod, shake, or tilt; idle is excluded.
The ground-truth intervals define the scored frames; they do not shift either method's predictions to reaction onset.

The baseline uses manually defined phrase dictionaries: agreement cues such as ``I agree'' and ``exactly'' vote for nod; negation or disagreement cues such as ``no'' and ``wrong'' vote for shake; and uncertainty or question cues such as ``not sure'' and ``what do you think'' vote for tilt.
Each user sentence is lowercased, and each dictionary entry contributes one vote if it appears with matching word boundaries; a question mark adds one tilt vote.
The class with the most votes is assigned throughout the sentence interval.
Ties follow nod, then shake, then tilt, with this priority selected on 299 validation dialogues and fixed before testing.
Frames with no matched cue or no active user sentence use shake, the most frequent reaction class in the training split.
The rules follow sentence intervals, so a reaction occurring after its triggering sentence is scored using the next active user sentence's decision or the default shake, without extending the preceding decision.
The baseline receives complete ground-truth user sentences and their timestamps, including words not yet spoken at a scored frame; gesture annotations are not prediction inputs.

The router selects the largest of its three reaction logits without decision confirmation.
The Nod, Shake, and Tilt rows report the fraction of correctly classified frames within each ground-truth class; Mean is their unweighted average.
These scores assess agreement with synthetic intended reaction types, not reaction detection or human-rated appropriateness.

\para{Router-History Ablation.}
Table~\ref{tab:ablation_semantic_routing} varies the number of current-inclusive hidden states pooled by the behavior router: 6, 13, 25, or 50 states at 12.5\,Hz, corresponding to 0.48, 1.04, 2.00, and 4.00\,s (nominally labeled 0.5, 1, 2, and 4\,s in the table).
These windows constrain the router's pooling interval, not the conversation context already encoded in each hidden state.
For this diagnostic, we freeze the final speech and motion models and shared speech projection, and retrain the reaction head for each history window.
Histories are zero-padded in hidden-state space at dialogue start.
Within each of three seeds, all four history conditions start from identical fresh head weights; the two-second head is retrained under the same protocol.
Using cached recorded-conversation hidden states and one motion variant per dialogue, we train on listening frames of 10,134 training dialogues for eight epochs with AdamW (learning rate $10^{-3}$, weight decay $10^{-4}$, batch size 4,096), using four-class cross-entropy weighted by inverse square-root training class frequencies.
Each head's checkpoint is selected by conditional three-class balanced accuracy on 299 validation dialogues before evaluating the 388-dialogue test set.
The scored set contains 38,375 annotated nod, tilt, or shake frames during listening; ground-truth reaction presence is supplied to every condition.
We exclude the idle logit and take the largest of the three reaction logits without decision confirmation.
Each class column reports recall averaged over three seeds; Mean reports the unweighted average over classes and its sample standard deviation across seeds.
This evaluates reaction-type agreement with synthetic intended labels, not reaction detection, timing, or generated-motion appropriateness.

\begin{table}[H]
    \centering
    \fontsize{8}{9}\selectfont
    \caption{Router history ablation (\%, $\uparrow$) on ground-truth reaction frames. Mean denotes balanced accuracy $\pm$ sample standard deviation over three seeds. Actual history windows are 0.48, 1.04, 2, and 4\,s; gray marks the default.}
    \label{tab:ablation_semantic_routing}
    \setlength{\tabcolsep}{3pt}
    \renewcommand{\arraystretch}{1.05}
\begin{tabularx}{\linewidth}{@{}l*{4}{>{\centering\arraybackslash}X}@{}}
    \toprule
    History (s) & Nod & Shake & Tilt & Mean \\
    \midrule
    0.5 & 72.10 & 76.63 & 82.19 & 76.97\,$\pm$\,0.41 \\
    1 & 74.04 & 78.03 & 84.43 & 78.83\,$\pm$\,0.36 \\
    \rowcolor{lightgray!40}
    2 & 77.00 & \textbf{80.23} & \textbf{85.04} & \textbf{80.76}\,$\pm$\,0.17 \\
    4 & \textbf{77.46} & 78.32 & 84.10 & 79.96\,$\pm$\,0.30 \\
    \bottomrule
\end{tabularx}

\end{table}

\subsection{Hidden-State Horizon and Window Placement}
\label{app:hidden_state_windows}

Table~\ref{tab:ablation_horizon} reports motion quality for horizons $\RunAheadHorizon=1,\ldots,6$ with windows of $\RunAheadHorizon+1$ hidden states.
Table~\ref{tab:ablation_window_position} varies the placement of a four-state window, with offsets $[\ell-3,\ell]$ relative to the target motion and one step equal to 80\,ms.
The $\RunAheadHorizon=3$ and $\ell=3$ rows retain the FacePlex reference from Table~\ref{tab:main}.
\begin{table}[H]
    \centering
    \fontsize{8}{9}\selectfont
    \caption{Motion quality for hidden-state horizons $\RunAheadHorizon=1,\ldots,6$, using windows of $\RunAheadHorizon+1$ states. All metrics $\downarrow$; gray marks the FacePlex reference.}
    \label{tab:ablation_horizon}
    \setlength{\tabcolsep}{3pt}
    \renewcommand{\arraystretch}{1.05}
\begin{tabularx}{\linewidth}{@{}c*{5}{>{\centering\arraybackslash}X}@{}}
    \toprule
    $\RunAheadHorizon$ & S-LVE & S-MHD & S-FDD & L-FDD & L-FID \\
    \midrule
    1 & 8.036 & 1.871 & 25.163 & 28.265 & 1.849 \\
    2 & 7.736 & 1.831 & 24.639 & 28.765 & 2.082 \\
    \rowcolor{lightgray!40}
    3 & \textbf{7.162} & \textbf{1.674} & \textbf{22.839} & \textbf{26.623} & \textbf{1.407} \\
    4 & 7.625 & 1.803 & 24.124 & 26.998 & 2.045 \\
    5 & 7.732 & 1.831 & 24.311 & 27.757 & 2.642 \\
    6 & 7.803 & 1.825 & 24.139 & 27.754 & 2.302 \\
    \bottomrule
\end{tabularx}

\end{table}

\begin{table}[H]
    \centering
    \fontsize{8}{9}\selectfont
    \caption{Placement of a window of four hidden states. Offsets are in 80\,ms steps relative to the target motion. All metrics $\downarrow$; gray marks the FacePlex reference.}
    \label{tab:ablation_window_position}
    \setlength{\tabcolsep}{3pt}
    \renewcommand{\arraystretch}{1.05}
\begin{tabularx}{\linewidth}{@{}lc*{5}{>{\centering\arraybackslash}X}@{}}
    \toprule
    $\ell$ & Offsets & S-LVE & S-MHD & S-FDD & L-FDD & L-FID \\
    \midrule
    0 & $[-3,0]$ & 8.869 & 2.023 & 27.405 & 29.372 & 1.717 \\
    1 & $[-2,+1]$ & 8.377 & 1.955 & 25.023 & 28.500 & 2.036 \\
    2 & $[-1,+2]$ & 7.871 & 1.854 & 24.192 & 28.052 & 1.894 \\
    \rowcolor{lightgray!40}
    3 (Ours) & $[0,+3]$ & \textbf{7.162} & \textbf{1.674} & \textbf{22.839} & \textbf{26.623} & \textbf{1.407} \\
    4 & $[+1,+4]$ & 7.296 & 1.694 & 24.563 & 27.860 & 1.973 \\
    5 & $[+2,+5]$ & 7.297 & 1.683 & 25.403 & 27.318 & 2.292 \\
    6 & $[+3,+6]$ & 7.343 & 1.725 & 23.930 & 26.978 & 2.728 \\
    \bottomrule
\end{tabularx}

\end{table}

\subsection{Hidden-State Timing Probe}
\label{app:hidden_state_probe}

For the analysis in Figure~\ref{fig:future_context_analysis}(a), we extract hidden states from the frozen speech backbone on Seamless Interaction~\citep{agrawal2025seamless} conversations.
We fit ridge regression models to predict motion at a target pair's position, each using a single hidden state at a specified temporal offset.
We compare past, current, and future positions relative to the target, with one step corresponding to 80\,ms.
The regressions are fitted on training data and evaluated on a separate test set.
We report the coefficient of determination $\ProbeRSquared$ for jaw, expression, head, and eye motion; higher values indicate more accurate predictions from the selected hidden state.
These scores characterize the predictive information available to a linear model at each offset.


\section{Moshi-Face Baseline Implementation}
\label{app:moshiface_implementation}

No public implementation of Moshi-Face was available at the time of our experiments, so we implement a baseline based on the method of \citet{jiang2026integrating}.
The Moshi-Face entry in Table~\ref{tab:main} refers to this implementation, with the adaptations below.

\para{Motion Representation and Codec.}
We use 108-dimensional FLAME parameters in place of the original vertex-based motion representation.
We retain the codec's temporal downsampling factor $r=2$, $N=8$ face tokens per downsampled timestep, codebook size $K=256$, and code embedding dimension $C=128$.
The codec uses a convolutional layer and a transformer layer and retains non-causal temporal processing.

\para{Face Transformer and Conditioning.}
The Face Transformer receives the conversational hidden state, learned positional embeddings, and previous face tokens, with non-causal attention across the $N$ token positions within a timestep.
We use only the conversational hidden state $h_i$ as its dialogue-conditioning vector, omitting the explicit text- and audio-token embedding sums used in the original method.
The hidden states retain contextual information from the speech model, but this conditioning differs from the original model's explicit token inputs.

\para{Training and Inference.}
We train on approximately 180 hours of real Seamless Interaction conversations.
For conversational-model adaptation, we implement only the first training stage: PersonaPlex remains frozen while the Face Transformer is trained using teacher forcing and cross-entropy loss on face tokens.
We train beyond the original 500-step schedule and select the checkpoint with the lowest held-out cross-entropy, rather than stopping at a fixed step count.
At inference, face tokens are sampled with temperature 0.8.


\section{User Study Details}
\label{app:user_study}

\subsection{Protocol and Participants}
\label{appsub:protocol}

We conducted a perceptual user study to complement the automatic metrics reported in the main paper.
A total of 31 participants were recruited to evaluate conversational speech-facial motion generation quality.
The study compared five methods: ARTalk, DualTalk, UniLS, Moshi-Face, and FacePlex.
Each participant evaluated 5 conversation sessions, and each session contained 5 videos generated from the same input conversation by the five methods.
In total, the study collected 775 video-level evaluations, corresponding to 155 ratings per method for each evaluation criterion.

The video materials were presented within the study interface using anonymized model labels, Model A--E.
Participants rated each video independently, and the model order was randomized for each session to reduce position bias.
Model identities were hidden from participants.

\subsection{Interface and Rating Criteria}
\label{appsub:interface}

\begin{figure}[t]
    \centering
    \includegraphics[width=\linewidth]{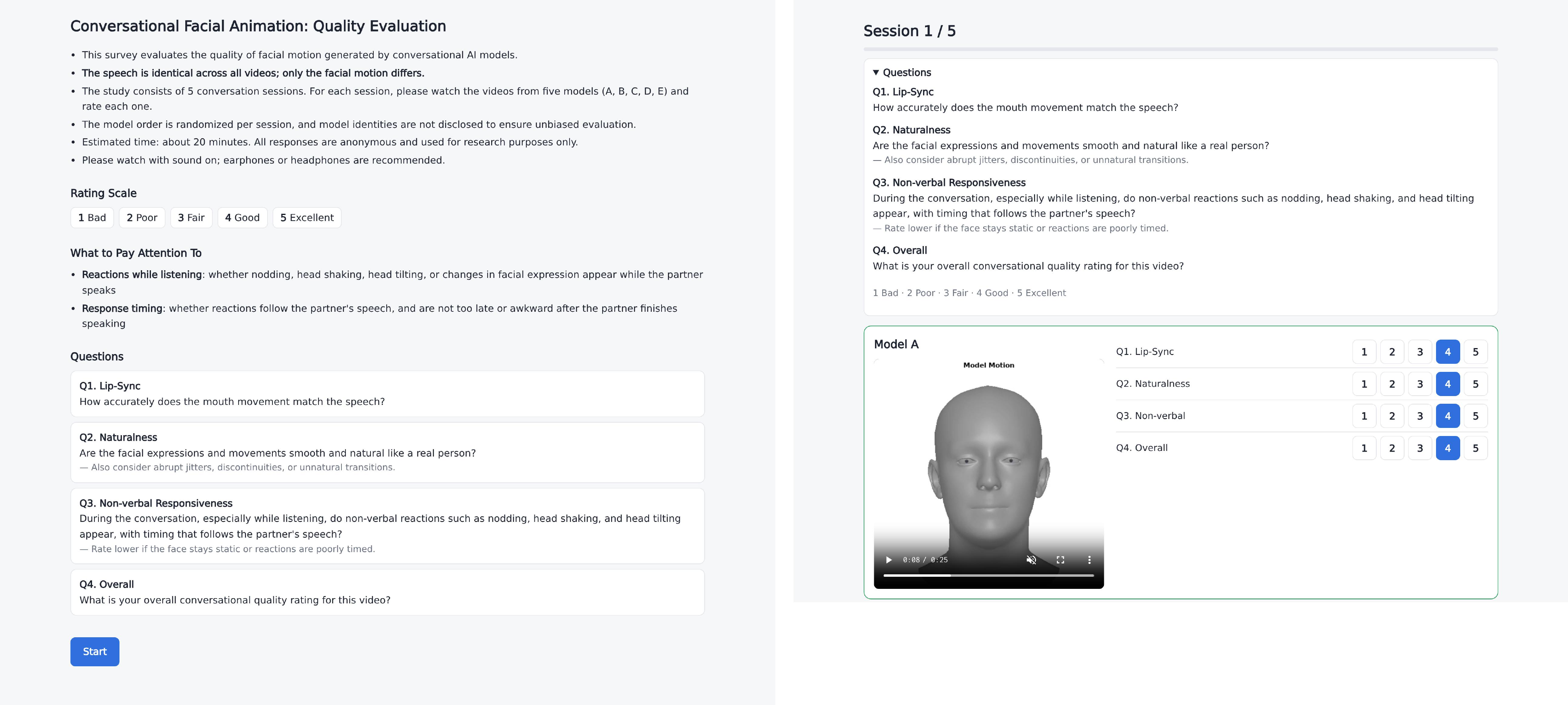}
    \caption{
    User study interface.
    Left: study instructions and the four evaluation criteria, rated on a 1--5 scale.
    Right: an example session with an anonymized model video and its rating controls.
    The four criteria are Lip-Sync, Naturalness, Non-verbal Responsiveness, and Overall conversational quality.
    Best viewed with zoom.
    }
    \label{fig:user_study_interface}
\end{figure}

The study used a web-based interface combining instructions, video playback, and rating controls, as shown in Figure~\ref{fig:user_study_interface}.
The left panel shows the instructions and evaluation criteria, and the right panel shows an example session with an anonymized model video and its rating controls.
Within each session, the speech was identical across model outputs, and only the facial motion differed.

Participants were instructed to watch each video with audio enabled; earphones or headphones were recommended.
The instructions described anonymous response collection and the 1--5 rating scale, with particular attention to reactions while listening and their timing relative to the partner's speech.
These reactions included nodding, head shaking, head tilting, and changes in facial expression.

Participants rated each video on a five-point scale: 1 (Bad), 2 (Poor), 3 (Fair), 4 (Good), and 5 (Excellent).
We report the mean opinion score for each criterion, abbreviated as Sync, Nat., NVR, and MOS in Table~\ref{tab:user}, respectively.
The questionnaire used the following four questions and accompanying instructions:
\begin{enumerate}[label=\textbf{Q\arabic*.},leftmargin=*,itemsep=0.4em]
    \item \textbf{Lip-Sync.} How accurately does the mouth movement match the speech?
    \item \textbf{Naturalness.} Are the facial expressions and movements smooth and natural like a real person?
    Also consider abrupt jitters, discontinuities, or unnatural transitions.
    \item \textbf{Non-Verbal Responsiveness.} During the conversation, especially while listening, do non-verbal reactions such as nodding, head shaking, and head tilting appear, with timing that follows the partner's speech?
    Rate lower if the face stays static or reactions are poorly timed.
    \item \textbf{Overall.} What is your overall conversational quality rating for this video?
\end{enumerate}

\section{End-to-End Processing Latency}
\label{app:latency}

\para{Processing Latency.}
We define end-to-end latency as the time from the onset of user speech to the first audible avatar response, consistent with Section~\ref{sec4.1}.
Table~\ref{tab:hardware_latency_components} reports speaking-interval means with FM compilation and CUDA Graphs enabled.
The H200 evaluation uses 22 sessions and two seeds, yielding 1,380 speaking ticks after warm-up and exclusion of the first 16 ticks per run; RTX 4090 results use a 120\,s live-server session.
Forward covers audio encoding, speech and motion generation, audio decoding, and streaming bookkeeping.
On H200, the reported latency to the first audible response is 122.16\,ms, comprising 80.00\,ms of input acquisition and 42.16\,ms of forward processing.
Rendering is reported separately per frame; mesh construction and rendering are excluded from Forward and Total.

\para{Future Audio-Token Readiness.}
Table~\ref{tab:ablation_run_ahead_latency} compares the three $k=3$ schedules on H200 using 22 sessions, two seeds, and 2,112 timed steps per condition after warm-up.
LM time measures synchronized computation from input codes on the GPU to the farthest output audio codes on the CPU, before Mimi decoding; codec and motion processing are excluded.
The total starts at the beginning of the current input interval and adds the required input wait to LM time.
For full-wait, earlier forward passes overlap input acquisition, so only the final pass follows the wait.

\section{Streaming-Optimized Avatar Rendering}
\label{app:optimization}

\subsection{Profiling Details}
\label{app:profiling}

We profile the renderer under the persistent low-batch streaming setting used by FacePlex. A substantial fraction of the avoidable latency originates not from the attention operator itself, but from fragmented execution around the transformer decoder. In particular, the operations between the Query/Key/Value projections and scaled dot-product attention are decomposed into many short GPU kernels. For a batch of two frames, Query/Key normalization and construction of the joint attention inputs require 31 kernel launches and 0.360\,ms, accounting for 33.5\% of the attention-module latency.

We therefore optimize the execution of the pretrained renderer without modifying its learned function. The deployed pipeline combines QKPack, joint frame-pair rendering,
frame-invariant StyleGAN modulation, frozen linear-weight pre-casting,
and fixed-shape CUDA Graph execution.

\subsection{QKPack: Stream-Preserving Attention Preparation}
\label{app:qkpack}

The FlexAvatar transformer processes two feature streams: an avatar representation and a motion-conditioning representation. Let
\[
Z_a \in \mathbb{R}^{B\times S_a\times D},
\qquad
Z_m \in \mathbb{R}^{B\times S_m\times D},
\]
where $S_a=1024$ and $S_m=5$ in our deployment. Following the MMDiT-style attention design~\citep{esser2024scaling}, each stream is independently projected into Query, Key, and Value features,
\[
(Q_r,K_r,V_r)=\operatorname{Proj}_r(Z_r),
\qquad r\in\{a,m\}.
\]

Query and Key normalization is performed separately for the two streams. Let $\gamma_r^Q$ and $\gamma_r^K$ denote the learned RMSNorm scales for stream $r$. For a per-head vector $x\in\mathbb{R}^{D_h}$, we define
\[
\mathcal{N}(x;\gamma)
=
\gamma\odot
\frac{x}
{\sqrt{\frac{1}{D_h}\sum_{j=1}^{D_h}x_j^2+\epsilon}}.
\]

Before the final precision and layout conversion, the reference attention inputs are
\[
\widetilde Q
=
\operatorname{Cat}
\left[
\mathcal{N}(Q_a;\gamma_a^Q),
\mathcal{N}(Q_m;\gamma_m^Q)
\right],
\]
\[
\widetilde K
=
\operatorname{Cat}
\left[
\mathcal{N}(K_a;\gamma_a^K),
\mathcal{N}(K_m;\gamma_m^K)
\right],
\qquad
\widetilde V
=
\operatorname{Cat}[V_a,V_m].
\]
Let $\mathcal{F}$ denote the deterministic precision conversion and layout transformation required by scaled dot-product attention. The final reference inputs are
\[
(Q^\star,K^\star,V^\star)
=
\mathcal{F}(\widetilde Q,\widetilde K,\widetilde V).
\]

The reference implementation realizes stream-specific Query/Key normalization, concatenation, precision conversion, and attention-input formatting as separate operations. Although individually lightweight, these operations introduce substantial launch overhead under low-batch streaming inference. QKPack replaces this sequence with a single fused Triton kernel that receives the same projected Query/Key/Value features and directly writes the final attention inputs while preserving the original stream-specific normalization parameters. Algorithms~\ref{alg:qkpack_ref} and~\ref{alg:qkpack} compare the two execution paths.

\begin{algorithm}[!t]
\caption{\textbf{Reference attention preparation}}
\label{alg:qkpack_ref}
\begin{algorithmic}[1]
\Statex \hspace{-\algorithmicindent}\textbf{Input:} $Q_a,K_a,V_a \in \mathbb{R}^{B\times H\times S_a\times D_h}$, $Q_m,K_m,V_m \in \mathbb{R}^{B\times H\times S_m\times D_h}$, and $\gamma_a^Q,\gamma_a^K,\gamma_m^Q,\gamma_m^K \in \mathbb{R}^{D_h}$
\Statex \hspace{-\algorithmicindent}\textbf{Output:} $Q^\star,K^\star,V^\star \in \mathbb{R}^{B\times H\times(S_a+S_m)\times D_h}$
\Statex
\State $\bar Q_a \gets \mathcal{N}(Q_a;\gamma_a^Q)$
\State $\bar K_a \gets \mathcal{N}(K_a;\gamma_a^K)$
\State $\bar Q_m \gets \mathcal{N}(Q_m;\gamma_m^Q)$
\State $\bar K_m \gets \mathcal{N}(K_m;\gamma_m^K)$
\State $\widetilde Q \gets \operatorname{Cat}(\bar Q_a,\bar Q_m)$
\State $\widetilde K \gets \operatorname{Cat}(\bar K_a,\bar K_m)$
\State $\widetilde V \gets \operatorname{Cat}(V_a,V_m)$
\State $(Q^\star,K^\star,V^\star) \gets \mathcal{F}(\widetilde Q,\widetilde K,\widetilde V)$
\end{algorithmic}
\end{algorithm}

\begin{algorithm}[!t]
\caption{\textbf{QKPack}}
\label{alg:qkpack}
\begin{algorithmic}[1]
\Statex \hspace{-\algorithmicindent}\textbf{Input:} Same Query/Key/Value features and RMSNorm scales as Alg.~\ref{alg:qkpack_ref}; $B$ batches, $H$ attention heads, $S_a$ avatar tokens, and $S_m$ motion-conditioning tokens
\Statex \hspace{-\algorithmicindent}\textbf{Output:} $Q^\star,K^\star,V^\star \in \mathbb{R}^{B\times H\times(S_a+S_m)\times D_h}$ directly in the attention layout
\Statex
\State $S \gets S_a+S_m$
\For{each batch $b$, attention head $h$, and joint-sequence position $s\in\{0,\ldots,S-1\}$ in parallel}
    \If{$s<S_a$} \Comment{avatar representation}
        \State $(q,k,v)\gets(Q_a,K_a,V_a)[b,h,s]$
        \State $(\gamma^Q,\gamma^K)\gets(\gamma_a^Q,\gamma_a^K)$
    \Else \Comment{motion condition}
        \State $s_m\gets s-S_a$
        \State $(q,k,v)\gets(Q_m,K_m,V_m)[b,h,s_m]$
        \State $(\gamma^Q,\gamma^K)\gets(\gamma_m^Q,\gamma_m^K)$
    \EndIf
    \State $\bar q\gets\mathcal{N}_{\mathrm{FP32}}(q;\gamma^Q)$
    \State $\bar k\gets\mathcal{N}_{\mathrm{FP32}}(k;\gamma^K)$
    \State $Q^\star[b,h,s]\gets\operatorname{Cast}(\bar q)$
    \State $K^\star[b,h,s]\gets\operatorname{Cast}(\bar k)$
    \State $V^\star[b,h,s]\gets v$
\EndFor
\end{algorithmic}
\end{algorithm}

Here, $b$, $h$, and $s$ index the batch element, attention head, and token position in the joint avatar--motion sequence, respectively. Each $(b,h,s)$ tuple is mapped to an independent Triton program instance and processed in parallel. The fixed boundary at $S_a$ determines whether the corresponding Query/Key/Value features and RMSNorm parameters are read from the avatar or motion-conditioning stream.

QKPack preserves the four independently learned normalization parameters,
\[
\gamma_a^Q,\qquad
\gamma_a^K,\qquad
\gamma_m^Q,\qquad
\gamma_m^K.
\]
To show equivalence to the reference execution, consider a joint-sequence position $s\in\{0,\ldots,S_a+S_m-1\}$. Before the final precision conversion,
\[
\widetilde Q_{\mathrm{QKPack}}[b,h,s]
=
\begin{cases}
\mathcal{N}\!\left(Q_a[b,h,s];\gamma_a^Q\right), & s<S_a,\\[2mm]
\mathcal{N}\!\left(Q_m[b,h,s-S_a];\gamma_m^Q\right), & s\ge S_a,
\end{cases}
\]
\[
\widetilde K_{\mathrm{QKPack}}[b,h,s]
=
\begin{cases}
\mathcal{N}\!\left(K_a[b,h,s];\gamma_a^K\right), & s<S_a,\\[2mm]
\mathcal{N}\!\left(K_m[b,h,s-S_a];\gamma_m^K\right), & s\ge S_a,
\end{cases}
\]
and
\[
\widetilde V_{\mathrm{QKPack}}[b,h,s]
=
\begin{cases}
V_a[b,h,s], & s<S_a,\\[1mm]
V_m[b,h,s-S_a], & s\ge S_a.
\end{cases}
\]
These expressions are exactly the element-wise definitions of the concatenated reference tensors $\widetilde Q$, $\widetilde K$, and $\widetilde V$. Since QKPack writes the same values directly into the positions defined by $\mathcal{F}$, in exact arithmetic,
\[
Q_{\mathrm{QKPack}}=Q^\star,
\qquad
K_{\mathrm{QKPack}}=K^\star,
\qquad
V_{\mathrm{QKPack}}=V^\star.
\]

For numerical stability, the RMS statistics are accumulated in \texttt{FP32}, consistent with the reference computation. QKPack then writes the normalized Query and Key features directly in the precision and layout consumed by the attention operator. Because the floating-point reduction schedule differs from the original sequence of PyTorch operations, QKPack is algorithmically equivalent but not bit-identical.

For a batch of two frames, the reference attention-preparation region requires 31 GPU kernel launches and 0.360\,ms. QKPack replaces this region with a single Triton kernel. Compared with the same execution path compiled using \texttt{torch.compile(max-autotune)}, QKPack further accelerates this region by $1.25\times$.

\subsection{Removing Frame-Invariant Computation}
\label{app:static}

Persistent streaming repeatedly evaluates the same frozen renderer while only the motion input changes. We therefore move computations that are invariant across streaming steps outside the per-frame execution.

\para{Frozen Linear-Weight Pre-Casting.}
The transformer linear weights are stored in \texttt{FP32}, while the relevant inference path is executed under \texttt{FP16} autocast. Since the renderer is frozen during deployment, the \texttt{FP32}$\rightarrow$\texttt{FP16} representation of each linear weight is identical across all streaming steps. Let $C_{16}$ denote the deterministic conversion to \texttt{FP16}. The reference execution repeatedly evaluates
\[
Y_t=X_t C_{16}(W)^\top.
\]
Instead, we compute
\[
\widetilde W=C_{16}(W)
\]
once during initialization and evaluate
\[
Y_t=X_t\widetilde W^\top
\]
during streaming. Because $W$ remains fixed,
\[
X_t C_{16}(W)^\top
=
X_t\widetilde W^\top,
\]
so the repeated conversion can be removed from the streaming loop without changing the resulting computation.

We apply this transformation only to the frozen linear layers inside the transformer path. Normalization parameters remain in \texttt{FP32}, and modules executed outside the autocast region are left unchanged. In our implementation, 160 linear weights are pre-cast. Across 576 real motion frames, frozen linear-weight pre-casting produces exactly the same five predicted Gaussian attributes as runtime casting, with maximum absolute difference equal to zero.

\para{Frame-Invariant StyleGAN Modulation.}
Style-modulated convolutions~\citep{karras2020analyzing} construct an effective convolution weight from a style vector at each layer. For layer $\ell$, let
\[
W_\ell
\in
\mathbb{R}^{C_{\mathrm{out}}\times C_{\mathrm{in}}\times K_h\times K_w}
\]
denote the convolution weight. We use $o$ and $i$ to index output and input channels, respectively, and $p$ to index a spatial position in the $K_h\times K_w$ convolution kernel. Let
\[
s_\ell=A_\ell(w)\in\mathbb{R}^{C_{\mathrm{in}}}
\]
denote the channel-wise modulation coefficients produced from style input $w$. The modulated convolution weight is
\[
W'_{\ell,oip}
=
W_{\ell,oip}\,s_{\ell,i}.
\]
StyleGAN demodulation rescales each output channel according to
\[
d_{\ell,o}
=
\left(
\sum_{i,p}
\left(W'_{\ell,oip}\right)^2
+\epsilon
\right)^{-1/2},
\]
yielding the effective convolution weight
\[
\widehat W_{\ell,oip}
=
d_{\ell,o}W'_{\ell,oip}.
\]

Whenever $w$ remains fixed during streaming, both $s_\ell$ and $\widehat W_\ell$ are independent of the current frame. The StyleGAN-based upsampling path used by FlexAvatar~\citep{kirschstein2026flexavatar} satisfies this condition in our inference configuration. We therefore construct $\widehat W_\ell$ once before streaming and subsequently execute an ordinary convolution with this fixed effective weight.

This transformation changes only when the effective convolution weight is constructed, not the convolution itself, and is theoretically bit-identical to runtime modulation when precision, arithmetic order, and kernel selection are held fixed. More generally, the same transformation applies to style-modulated rendering blocks whenever their style input remains fixed throughout streaming.

\subsection{Fixed-Shape CUDA Graph Execution}
\label{app:cudagraph}

Under the fixed input shapes used for streaming deployment, the renderer follows a repeatable GPU execution sequence across steps. We capture this sequence with CUDA Graphs and replay it at subsequent steps. In the final deployed configuration, QKPack and the frame-invariant transformations further reduce the work contained in the captured path.

\subsection{Runtime Ablation}
\label{app:runtime_ablation}

Table~\ref{tab:renderer_runtime} reports cumulative deployment configurations rather than isolated operator-level effects. For the two frames generated at each 80-ms FacePlex streaming step, the original and final configurations require 13.230 and 4.052\,ms, respectively, corresponding to 16.5\% and 5.1\% of the streaming interval. Kernel-level profiling further shows that the number of CUDA kernels decreases from 51 to 12 in the profiled attention path and from 83 to 36 in the complete transformer block.

\subsection{Numerical Equivalence and Rendering Quality}
\label{app:renderer_equivalence}

The deployed optimizations modify execution rather than the learned rendering function. We call a transformation \emph{theoretically bit-identical} when it preserves the numerical operations with precision, arithmetic order, and kernel selection held fixed. Algorithmically equivalent transformations preserve the mathematical computation but can change floating-point rounding through a different execution order. These properties describe individual transformations; Table~\ref{tab:renderer_runtime} measures rendered-image differences between complete deployment configurations and the original renderer.

\begin{table}[t]
\centering
\caption{Theoretical numerical properties of individual renderer optimizations under fixed precision and kernel selection.}
\label{tab:renderer_exactness}
\resizebox{0.9\linewidth}{!}{
\begin{tabular}{lc}
\toprule
\textbf{Optimization} & \textbf{Numerical property} \\
\midrule
CUDA Graph
    & Theoretically bit-identical \\
Joint frame-pair rendering
    & Algorithmically equivalent, non-bit-identical \\
QKPack
    & Algorithmically equivalent, non-bit-identical \\
Frame-invariant StyleGAN modulation
    & Theoretically bit-identical \\
Frozen linear-weight pre-casting
    & Theoretically bit-identical to runtime casting \\
\bottomrule
\end{tabular}}
\end{table}

CUDA Graph capture, frame-invariant StyleGAN modulation, and frozen linear-weight pre-casting are theoretically bit-identical under the conditions above. Joint frame-pair rendering and QKPack are algorithmically equivalent to their reference executions but not bit-identical: batching the two frames can alter the GPU kernel execution path, while QKPack changes the floating-point execution order through fused normalization and packing. Across the complete optimized pipeline, the resulting numerical differences remain perceptually negligible, with LPIPS $1.04\times10^{-5}$ relative to the original renderer.

We do not use quantization, pruning, low-rank approximation, temporal frame reuse, or model fine-tuning. The reported acceleration therefore results from reorganizing the execution of the pretrained renderer rather than approximating or retraining its rendering function.

\section{Limitations and Broader Impact}
\label{app:limitations}

\subsection{Limitations}
\label{appsub:limitations}

Although FacePlex takes a step toward full-duplex joint speech--facial motion generation, several limitations remain.

First, our motion-quality benchmarks primarily evaluate FLAME-based geometry and motion dynamics.
Our renderer comparisons assess numerical fidelity to the original renderer, but do not comprehensively evaluate rendering artifacts, identity preservation, or robustness to changes in lighting and appearance.
In addition, conversational facial behavior, especially listener motion, is inherently one-to-many: multiple reactions may be plausible for the same dialogue context.
Thus, geometric and distributional metrics such as LVE, MHD, FDD, and L-FID only partially capture conversational appropriateness, affective nuance, and semantic grounding.
We complement these metrics with a perceptual user study, but broader evaluations across participants, languages, identities, and conversational contexts remain necessary.
Moreover, the main motion-quality comparison uses recorded conversations with future user audio available; these results do not directly establish motion quality under causal run-ahead conditioning.

Second, part of our training data combines language-model-generated dialogues, synthesized speech, and teacher-generated FLAME motion augmented with gesture primitives.
This provides scalable supervision, but may inherit biases, failure modes, and limited motion diversity from the language model, motion teacher, and composition rules.
The synthetic dialogues contain no overlapping speech, and the explicit reaction vocabulary is limited to idle, nod, shake, and tilt.
FacePlex models facial and head motion during both speaking and listening, while body gestures, hand motion, and visual scene context remain outside its scope.
Although fixed persona tokens condition the avatar's motion characteristics, consistency over prolonged interactions has not been systematically evaluated.
Improving data diversity, expanding the reaction vocabulary, and evaluating longer and more varied interactions are important directions for future work.

\subsubsection{Failure Case Analysis}
\label{appsub:failure_cases}

\begin{figure}[t!]
    \centering
    \begingroup
    \newcommand{\failurepanel}[2]{%
        {\small #1\par}\smallskip
        \begin{tikzpicture}
            \node[inner sep=0] (panel) {\includegraphics[width=0.90\linewidth]{#2}};
            \path (panel.north west) -- (panel.south west)
                node[pos=0.25,left=5pt,font=\small\bfseries] {GT}
                node[pos=0.75,left=5pt,font=\small\bfseries] {Ours};
        \end{tikzpicture}\par
    }
    \failurepanel{(a) \textbf{Lip under-articulation} (``popsicle'')}{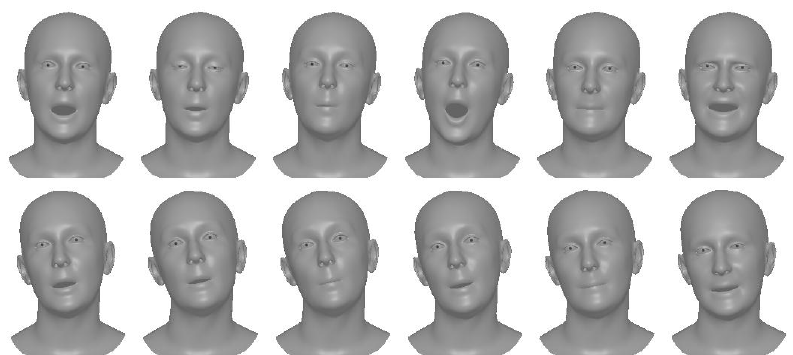}
    \medskip
    \failurepanel{(b) \textbf{Head over-rotation} (listening)}{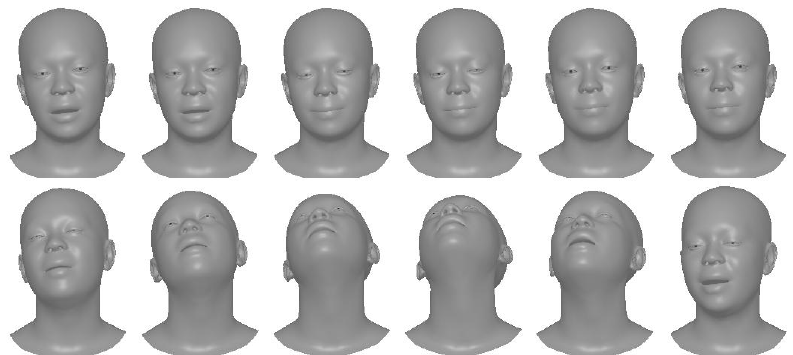}
    \medskip
    \failurepanel{(c) \textbf{Excessive head turn and mouth opening} (listening)}{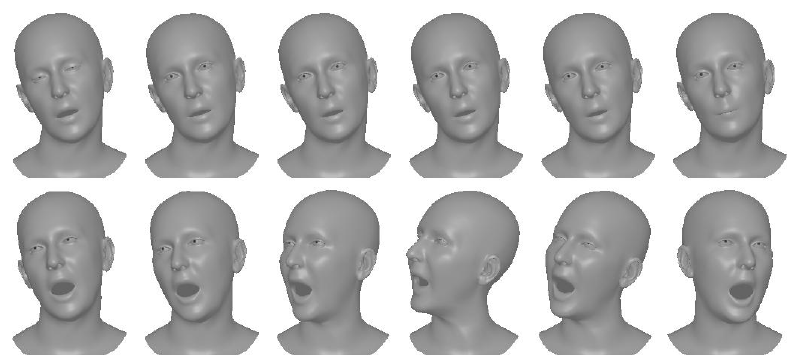}
    \endgroup
    \caption{\textbf{Failure cases.}
    (a) On the stressed vowel of ``popsicle'', Ours under-opens the mouth (lip extent 36\,mm vs.\ 56\,mm in GT).
    (b) While listening, Ours raises the head by up to $56^{\circ}$ although the GT stays frontal.
    (c) While listening, Ours turns the head by up to $50^{\circ}$ and opens the mouth, while the GT remains nearly still.
    These cases come from the most severe deviations among the evaluated clips from the Seamless Interaction test split.}
    \label{fig:failure_cases}
\end{figure}

Figure~\ref{fig:failure_cases} illustrates three failure cases selected from the most severe deviations among the evaluated clips from the Seamless Interaction test split, with GT in the upper row and FacePlex (Ours) in the lower row of each panel.
The speaking example shows insufficient mouth opening on a stressed vowel, while the two listening examples show excessive head rotation and unintended mouth opening relative to GT.
These examples highlight remaining limitations in articulatory amplitude during speech and in controlling head and mouth motion while listening.

\subsection{Potential Positive Impact}
\label{appsub:pos_impact}

FacePlex may contribute to more natural and accessible real-time conversational interfaces.
By coordinating speech with facial articulation and dialogue-grounded reactions during speaking and listening, the framework could support applications in telepresence, remote education, language learning, entertainment, and assistive communication.
Such interfaces may be particularly useful to users who rely on visual speech cues, facial expressions, and response timing, although these benefits require evaluation with the intended user populations.

FacePlex also provides a framework for studying online multimodal interaction.
It coordinates speech generation, facial motion, and rendering on a shared streaming timeline, using predicted upcoming speech and updated conversational hidden states to condition motion.
This approach may support further research on anticipatory articulation and context-dependent non-verbal behavior under streaming constraints.

\subsection{Potential Negative Impact}
\label{appsub:neg_impact}

The same capabilities that make full-duplex conversational avatars more natural also introduce potential risks.
Speech-synchronized facial motion and realistic avatar rendering could be misused to create deceptive synthetic media, impersonate real individuals, or generate avatars without proper consent.
When combined with voice cloning or used to reproduce an identifiable person's appearance, such systems could facilitate social engineering, misinformation, or unauthorized digital replicas.

There are also privacy, consent, and overtrust concerns.
Conversational-avatar systems may use speech, motion, video, or identity-related data for training or personalization.
Inadequate consent and data protection could expose sensitive information or enable unauthorized reproduction of a person's voice or appearance.
Moreover, responsive speech and facial behavior may lead users to attribute more understanding, empathy, or reliability to the system than is warranted, particularly in high-stakes settings such as healthcare, education, hiring, counseling, or legal assistance.

To mitigate these risks, deployment should include clear disclosure that the avatar is synthetic, appropriate consent and data protection practices, usage restrictions, and safeguards against impersonation or non-consensual identity use.
Watermarking or provenance mechanisms may provide additional support where applicable.
Our work is intended for research on streaming multimodal generation and should not be used to create deceptive or unauthorized synthetic representations of real people.

\end{document}